\documentclass[11pt]{article}

\usepackage[final]{acl}
\usepackage{times}
\usepackage{latexsym}

\usepackage[T1]{fontenc}
\usepackage[utf8]{inputenc}

\usepackage{microtype}

\usepackage{inconsolata}

\usepackage{graphicx}
\usepackage{booktabs}
\usepackage{amsmath, amssymb}
\usepackage{tikz}
\usetikzlibrary{arrows.meta, positioning, calc, fit, backgrounds, shapes.geometric}
\usepackage{pgfplots}
\usepgfplotslibrary{groupplots}
\pgfplotsset{compat=1.18}

\usepackage{multirow}
\usepackage{arydshln}
\usepackage{makecell}
\usepackage{graphicx}
\title{EDWARD-NMT: Elastic Dimensions With Adaptive Routing and Dual-consistency for Neural Machine Translation}

\author{
 \textbf{Edward Nguyen}\thanks{Equal contribution.},
 \textbf{Thai Tran}\footnotemark[1],
 \textbf{Kiet Anh Hoang}\footnotemark[1],
 \textbf{Duy Hoang}\footnotemark[1],
 \textbf{Duy Vo},
 \textbf{Khanh Le},
 \textbf{Vu Le}\\
\\
UNEY, Switzerland
\\
\texttt{\{edward.nguyen, thai.tt, kiet.ha, duy.hv, duy.vhb, khanh.ld, vu.lt\}@uney.com}
}

\begin{document}
\maketitle
\begin{abstract}

Sparse Mixture-of-Experts (MoE) is a powerful paradigm for scaling Neural Machine Translation (NMT) and managing language heterogeneity. However, practical limitations hinder its deployment on resource-constrained devices: the entire expert pool must remain in memory, and the per-token compute budget is statically fixed during training by rigid top-$k$ routing. We introduce \textbf{EDWARD-NMT} (\textbf{E}lastic \textbf{D}imensions \textbf{W}ith \textbf{A}daptive \textbf{R}outing and \textbf{D}ual-consistency for \textbf{NMT}), a novel MoE encoder-decoder architecture designed for a ``train once, deploy anywhere'' paradigm that exposes a family of deployable sub-networks at inference time. EDWARD-NMT integrates three key innovations: (i) Matryoshka-style elastic dimensions to flexibly reduce the memory footprint; (ii) an adaptive sigmoid router with a learned, task-conditioned threshold, allowing each token to dynamically determine its active expert count; and (iii) a dual-consistency objective to maximize the performance of the extracted sub-networks. Extensive experiments on PhoMT (En$\leftrightarrow$Vi), WMT-14 (En$\leftrightarrow$De), multilingual OPUS-16, and a multi-domain De$\leftrightarrow$En benchmark validate our approach. The full-capacity architecture matches or exceeds strong MoE baselines with improved parameter efficiency. Furthermore, the sub-networks seamlessly extracted from a single checkpoint consistently outperform comparably sized dense Transformers without any retraining.

% Furthermore, the smallest sub-network, EDWARD-NMT-S, outperforms a similarly sized Transformer Base by $+0.8$/$+0.7$ BLEU on PhoMT and $+1.9$/$+2.3$ BLEU on WMT-14, with all variants seamlessly extracted from a single checkpoint without retraining.

% Extensive experiments on PhoMT (En$\leftrightarrow$Vi), WMT-14 (En$\leftrightarrow$De), and a five-domain De$\leftrightarrow$En benchmark demonstrate the efficacy of our approach. The full-capacity EDWARD-NMT-XL improves upon a Transformer Big baseline by up to $+1.30$ BLEU while activating fewer parameters. Furthermore, the smallest sub-network, EDWARD-NMT-S, outperforms a similarly sized Transformer Base by $+0.8$/$+0.7$ BLEU on PhoMT and $+1.9$/$+2.3$ BLEU on WMT-14, with all variants seamlessly extracted from a single checkpoint without retraining.
\end{abstract}

\newcommand{\midsize}{\fontsize{8}{9}\selectfont}

\section{Introduction}
\label{sec:Intro}

Managing language and domain heterogeneity is a fundamental challenge in Neural Machine Translation (NMT) \cite{britz2017effective, koehn2017six, pham2021revisiting}, as training corpora routinely mix diverse lexical and stylistic distributions \cite{stahlberg2020neural, gain2026bridging}. The Mixture-of-Experts (MoE) paradigm addresses this by routing tokens to conditionally activated experts, scaling capacity to specialize in distinct patterns while keeping per-token computation tractable \cite{shazeer2017outrageously, fedus2022switch, dai2022stablemoe, zhang2024lightweight, liang2025thor}. 

Despite state-of-the-art performance, MoEs face deployment bottlenecks in resource-constrained environments. First, the entire expert pool must reside in memory \cite{hoffmann2022empirical, xu2023towards}. While lightweight adapters reduce this overhead \cite{zhang2024lightweight}, they still yield static, hardware-specific models. Second, standard MoEs rely on rigid top-$k$ routing, applying an identical compute budget regardless of token complexity \cite{jawahar2023automoe, guo2025dynamic}. Although Neural Architecture Search can design heterogeneous sub-networks \cite{jawahar2023automoe}, it requires complex offline phases. Consequently, NMT practitioners lack a unified framework to dynamically adjust memory and compute footprints on the fly.

To address these gaps, we propose \textbf{EDWARD-NMT} (\textbf{E}lastic \textbf{D}imensions \textbf{W}ith \textbf{A}daptive \textbf{R}outing and \textbf{D}ual-consistency for \textbf{NMT}), an encoder-decoder architecture enabling a \emph{train-once, deploy-anywhere} paradigm through two coordinated forms of adaptivity. To overcome memory bottlenecks, we apply Matryoshka representation learning \cite{kusupati2022matryoshka} to expert dimensions, allowing the MoE to be deterministically sliced into weight-sharing sub-networks at inference. To eliminate rigid compute constraints, we replace top-$k$ selection with an adaptive sigmoid router using a learned, task-conditioned threshold, empowering tokens to dynamically dictate their active expert count. Finally, to ensure aggressively truncated sub-networks maintain translation quality without post-hoc fine-tuning, we introduce a novel in-place \emph{dual-consistency} objective that explicitly aligns predictive distributions across nested structural scales.

% we introduce a novel in-place \emph{dual-consistency} objective that aligns predictive distributions across dimensional scales.

% Extensive experiments on PhoMT (En$\leftrightarrow$Vi)~\cite{doan2021phomt}, WMT-14 (En$\leftrightarrow$De)~\cite{bojar2014findings}, multilingual OPUS-16~\cite{zhao2024sparse-lingual-smoe}, and a multi-domain De$\leftrightarrow$En~\cite{aharoni2020unsupervised} benchmark validate our approach. EDWARD-NMT achieves strong full-capacity performance, improving upon Transformer Big~\cite{vaswani2017attention} by $+0.50$ to $+1.48$ BLEU across benchmarks, while keeping smaller variants highly capable. Our full model matches recent lightweight MoEs \cite{zhang2024lightweight}, while our smallest variant (EDWARD-NMT-S) outperforms a dense Transformer Base by $+0.49$ to $+3.18$ BLEU, with all variants seamlessly extracted from a single checkpoint.
Extensive experiments on PhoMT (En$\leftrightarrow$Vi)~\cite{doan2021phomt}, WMT-14 (En$\leftrightarrow$De)~\cite{bojar2014findings}, multilingual OPUS-16~\cite{zhao2024sparse-lingual-smoe}, and a multi-domain De$\leftrightarrow$En~\cite{aharoni2020unsupervised} benchmark validate our approach. EDWARD-NMT achieves strong full-capacity performance, improving upon Transformer Big~\cite{vaswani2017attention} across benchmarks, while keeping smaller variants highly capable. Our full model matches recent lightweight MoEs~\cite{zhang2024lightweight}, while our smallest variant (EDWARD-NMT-S) outperforms a dense Transformer Base, with all variants seamlessly extracted from a single checkpoint.
In summary, our main contributions are:
\begin{itemize}
\item We propose \textbf{EDWARD-NMT}, coupling elastic expert dimensions with dynamic, token-level sigmoid routing to realize a \emph{train-once, deploy-anywhere} NMT architecture.
% \item We introduce a \emph{dual-consistency} self-distillation objective that unifies structural regularization and in-place alignment, preserving the translation quality of aggressively truncated sub-networks.
\item We formulate a \emph{dual-consistency} self-distillation objective, an entirely in-place regularization mechanism that stabilizes aggressively sliced sub-networks without requiring post-hoc adaptation or fine-tuning.
\item We demonstrate that a single training run yields a discrete family of high-performing sub-networks, with the full-capacity model matching strong MoE baselines and smaller variants outperforming comparably sized dense networks.

\end{itemize}

\section{Related Work}
\label{sec:related}

\paragraph{Static vs. Elastic Mixture-of-Experts.}
Sparse MoEs dominate sequence model scaling \cite{shazeer2017outrageously, fedus2022switch, dai2022stablemoe, jiang2024mixtral, liu2024deepseek} and multi-domain NMT \cite{costa2022no, zhang2024lightweight}, yet produce static models with fixed memory footprints. Post-hoc compression \cite{lagunas2021block, li2025slimmoe} yields rigid, hardware-specific models, while multi-sub-network architectures \cite{cai2020once, yu2019slimmable, hou2020dynabert} require explicit fine-tuning. Existing elastic designs either restrict flexibility to dense networks \cite{kudugunta2024matformer} or limit it to token-level granularity \cite{cappellazzo2026mome}. While concurrent work (MoSE \cite{tastan2026mose}) introduces slimmable experts, it retains rigid top-$k$ routing and employs test-time training for width adaptation. In contrast, EDWARD-NMT embeds Matryoshka-style elasticity directly into sparse experts. By introducing a variance-preservation scaling factor, we stabilize activations across nested widths, allowing a single trained MoE to yield a family of deployable sub-networks without any post-hoc adaptation.

% \paragraph{Dynamic Routing and Adaptive Compute.}
% Traditional MoEs use rigid top-$k$ routing, applying identical compute budgets regardless of token complexity and relying on restrictive load-balancing losses~\cite{clark2022unified}. Efforts to soften this include soft \cite{puigcerver2024sparse} and expert-choice routing \cite{zhou2022mixture}. Recent work addresses per-token compute directly: AdaMoE \cite{zeng2024adamoe} introduces null experts, and DynMoE \cite{guo2025dynamic} auto-tunes activations during training. Most related is ReMoE \cite{wang2025remoe}, using a differentiable ReLU router to vary active experts. EDWARD-NMT is concurrent but distinct: by employing a sigmoid router with a task-conditioned threshold, coupled with elastic experts, our framework uniquely allows dynamic adjustment of both memory and compute footprints at inference.

\paragraph{Dynamic Routing and Adaptive Compute.}
Traditional MoEs employ rigid top-$k$ routing, applying identical compute budgets regardless of token complexity and relying on restrictive load-balancing losses~\cite{clark2022unified}. Relaxations include soft~\cite{puigcerver2024sparse} and expert-choice routing~\cite{zhou2022mixture}, while recent methods adapt per-token compute dynamically via null experts (AdaMoE~\cite{zeng2024adamoe}), auto-tuned activations (DynMoE~\cite{guo2025dynamic}), or differentiable ReLU routing (ReMoE~\cite{wang2025remoe}). EDWARD-NMT is concurrent but distinct: by coupling elastic experts with an adaptive sigmoid router, our framework uniquely enables dynamic adjustment of both memory and compute footprints at inference.

% \paragraph{Self-distillation and Consistency Regularization.}
% To enhance model generalization and robustness, typical consistency regularization aligns predictions across perturbed forward passes or augmented data views \cite{xie2020unsupervised, wu2021r, yao2025cr}, while self-distillation transfers knowledge between layers or temporal states \cite{furlanello2018born, zhu2018knowledge, mobahi2020self}. Although in-place distillation successfully trains dense elastic networks by aligning sub-networks with a full-capacity teacher \cite{yu2019universally, cai2020once}, it remains confined to statically routed architectures. We extend this via a novel \emph{dual-consistency objective} tailored for elastic MoEs. Rather than aligning augmented views, we dynamically transfer knowledge between structural widths of the \emph{same} network using a symmetric Kullback-Leibler (Jeffreys) divergence. This bidirectional regularization anchors the full-capacity network while preserving the quality of aggressively truncated variants.

\paragraph{Self-distillation and Consistency Regularization.}
To enhance model generalization and robustness, typical consistency regularization aligns predictions across perturbed forward passes or augmented data views \cite{xie2020unsupervised, wu2021r, yao2025cr}, while self-distillation transfers knowledge between layers or temporal states \cite{furlanello2018born, zhu2018knowledge, mobahi2020self}. Although in-place distillation successfully trains dense elastic networks by aligning sub-networks with a full-capacity teacher \cite{yu2019universally, cai2020once}, it remains confined to statically routed architectures. We extend this via a novel \emph{dual-consistency objective} tailored for elastic MoEs. Rather than aligning augmented views, we dynamically transfer knowledge between nested widths using a symmetric Kullback-Leibler (Jeffreys) divergence. This bidirectional regularization anchors the full-capacity network while preserving the quality of aggressively truncated variants.

\begin{figure*}[!ht]
  \centering
  \includegraphics[width=0.98\linewidth]{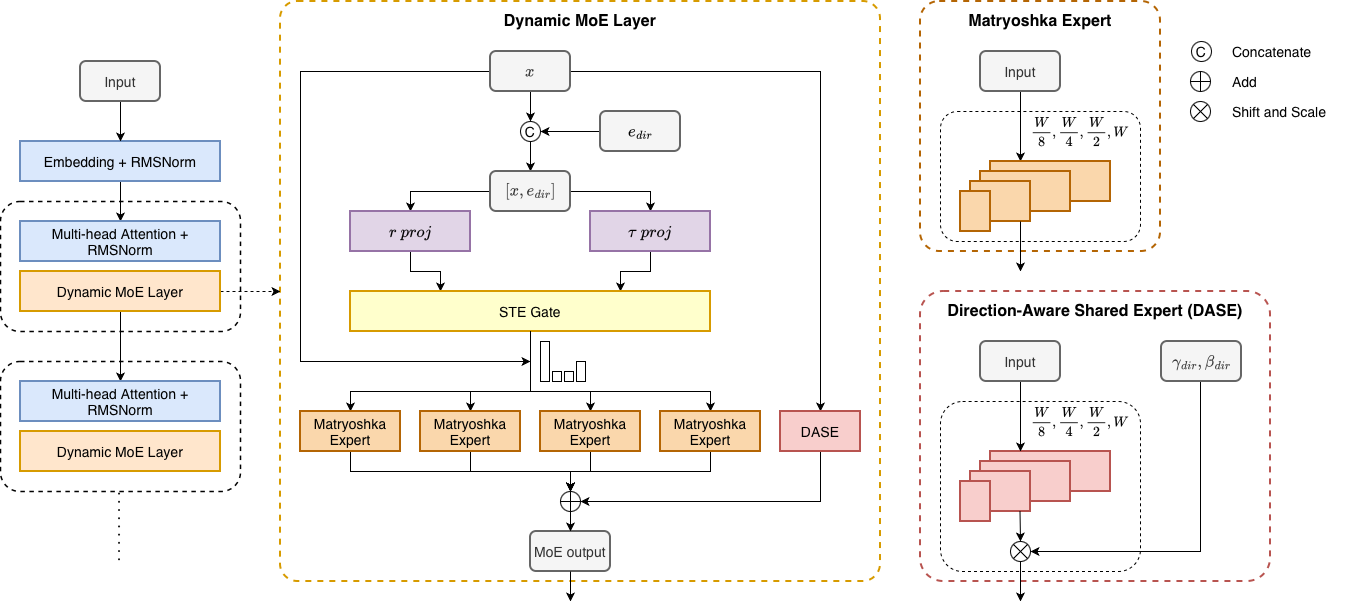}
  % \caption{Overview of the \textbf{EDWARD-NMT} architecture. \textbf{Left:} The macroscopic Transformer block integrating the Dynamic MoE layer. \textbf{Center:} The Adaptive Sigmoid Router utilizes a direction embedding ($e_{dir}$) to compute a token-specific threshold ($\tau$) and dynamically allocate the sparse expert budget. \textbf{Top Right:} The internal structure of a Matryoshka Expert, illustrating the nested parameter slices used to extract varying capacities. \textbf{Bottom Right:} The Direction-Aware Shared Expert (DASE) applies FiLM modulation ($\gamma_{dir}, \beta_{dir}$) to provide a stable, direction-aware baseline.}
  \caption{Overview of the \textbf{EDWARD-NMT} architecture. \textbf{Left:} The macro-level Transformer block integrating the Dynamic MoE layer. \textbf{Center:} The Adaptive Sigmoid Router computes a token-specific threshold ($\tau$) conditioned on a translation direction embedding ($e_{dir}$) to dynamically allocate the sparse expert budget. \textbf{Top Right:} The internal structure of a Matryoshka Expert, illustrating the nested parameter slices used to extract varying capacities. \textbf{Bottom Right:} The Direction-Aware Shared Expert (DASE) applies FiLM modulation ($\gamma_{dir}, \beta_{dir}$) to provide a stable, direction-aware backbone.}
  \label{fig:edward_nmt}
\end{figure*}

% \paragraph{Shared Experts and Knowledge Transfer.}
% Recent architectures like DeepSeek-MoE~\cite{dai2024deepseekmoe} and Qwen3~\cite{yang2025qwen3} isolate common knowledge into a dense, shared expert evaluated for every token, improving parameter efficiency. EDWARD-NMT builds upon this by dynamically conditioning the shared expert on the target translation direction via FiLM modulation \cite{perez2018film}, creating a unified, task-aware backbone without the computational overhead of dedicated, task-specific experts.

% \paragraph{Shared Experts and Knowledge Transfer.}
% To mitigate sparse experts from redundantly learning general linguistic patterns, architectures like DeepSeek-MoE~\cite{dai2024deepseekmoe} and Qwen3~\cite{yang2025qwen3} isolate common knowledge into a dense, universally evaluated shared expert. EDWARD-NMT extends this by dynamically conditioning the shared expert on the translation direction via FiLM modulation~\cite{perez2018film}, creating a unified, direction-aware backbone without the parameter overhead of dedicated experts.

% \paragraph{Shared Experts.}
% To prevent sparse experts from redundantly learning general linguistic patterns, architectures like DeepSeek-MoE~\cite{dai2024deepseekmoe} and Qwen3~\cite{yang2025qwen3} isolate common knowledge into a dense, universally evaluated shared expert. EDWARD-NMT extends this by dynamically conditioning the shared expert on the translation direction via FiLM modulation~\cite{perez2018film}, creating a unified, direction-aware backbone without the parameter overhead of dedicated experts.

\paragraph{Shared Experts.}
To prevent sparse experts from redundantly learning general linguistic patterns, architectures like DeepSeek-MoE~\cite{dai2024deepseekmoe} and Qwen3~\cite{yang2025qwen3} isolate common knowledge into a dense, universally shared expert. EDWARD-NMT extends this by dynamically conditioning the shared expert on the translation direction via FiLM modulation~\cite{perez2018film}, creating a unified, direction-aware backbone without the parameter overhead of dedicated experts.

\section{EDWARD-NMT}
\label{sec:method}

EDWARD-NMT builds upon the Transformer encoder-decoder architecture \cite{vaswani2017attention, shazeer2020glu}, selected over decoder-only alternatives for its superior parameter efficiency and low inference latency, making it highly practical for resource-constrained deployment. We replace conventional feed-forward sublayers with a novel MoE block comprising three jointly optimized innovations: (i) \textbf{elastic experts} encoding a family of nested inner widths, (ii) an \textbf{adaptive sigmoid router} dynamically allocating the optimal expert count per token, and (iii) a \textbf{dual-consistency} objective aligning predictions across capacity views. Additionally, a lightweight \textbf{Direction-Aware Shared Expert (DASE)} is computed in parallel. Figure~\ref{fig:edward_nmt} provides a schematic overview.

\subsection{Preliminaries: Standard Mixture-of-Experts}
\label{sec:method:prelim}

We adopt the sparsely gated MoE feed-forward layer \cite{shazeer2017outrageously, fedus2022switch} as our baseline. Let $x \in \mathbb{R}^{d_{\text{model}}}$ denote the input token representation. An MoE layer consists of $E$ independent expert networks, $\{E_i\}_{i=1}^E$, and a routing function $g_{\theta,k}(x)$ parameterized by a weight matrix $W_g \in \mathbb{R}^{d_{\text{model}} \times E}$.

The standard router projects the input to unnormalized logits $z = x W_g$. The routing weights are thus computed as $g_{\theta,k}(x) = \mathrm{Softmax}(\mathrm{TopK}(x W_g, k))$, and the MoE layer outputs their weighted sum:
\begin{equation}
\text{MoE}(x) = \sum_{i \in \mathcal{T}_k(x)} g_{\theta,k}(x)_i \cdot E_i(x),
\end{equation}
where $\mathcal{T}_k(x)$ is the set of selected expert indices. Each expert $E_i$ is typically a SwiGLU Multi-Layer Perceptron \cite{shazeer2020glu}, parameterized by up-projection weights $W_1, W_2 \in \mathbb{R}^{d_{\text{model}} \times M}$ and down-projection weights $W_3 \in \mathbb{R}^{M \times d_{\text{model}}}$, where $M$ is the maximum inner dimension.

This standard formulation exhibits two critical limitations: rigid top-$k$ routing fixes the per-token compute budget, and the physical memory footprint scales linearly with $E \cdot M$ regardless of how many experts are actively evaluated.

\subsection{Elastic Experts via Matryoshka Dimensions}
\label{sec:method:elastic}

Inspired by Matryoshka Representation Learning \cite{kusupati2022matryoshka, kudugunta2024matformer}, we design experts to support a nested family of inner dimensions $\mathcal{D} \subseteq \{1, \dots, M\}$ sharing a \emph{single} set of parameter tensors. During each forward pass, a target dimension $D_k \in \mathcal{D}$ is sampled. Rather than instantiating physically separate experts, we dynamically truncate intermediate activations to their first $D_k$ components. Letting $[\,:D_k]$ denote the hidden dimension slice, the elastic SwiGLU expert computes:
\begin{align}
h &= \mathrm{SiLU}(xW_1) \odot (xW_2), \\
E_i(x; D_k) &= \psi(D_k) \cdot \bigl(h_{[:D_k]} W_{3, [:D_k, :]}\bigr).
\end{align}

\paragraph{Variance Preservation.} 
Truncating the inner dimension intrinsically shrinks the magnitude of the down-projected activations, destabilizing residual additions and RMSNorm statistics \cite{zhang2019root}. To compensate, we introduce a deterministic rescaling factor $\psi(D_k)$ dependent strictly on the chosen width:
\begin{equation}
\psi(D_k) = \sqrt{\frac{M}{\max(D_k, 1)}}.
\end{equation}
This ensures the expected squared norm of $E_i(x; D_k)$ remains invariant to $D_k$, a correction paramount for stable joint training across all nested capacities. Consequently, a single trained network seamlessly exposes $|\mathcal{D}|$ weight-sharing sub-networks freely selectable at inference.

\subsection{Adaptive Sigmoid Routing}
\label{sec:method:routing}

To decouple the model from rigid top-$k$ constraints, we replace the standard softmax gate with a token-adaptive \emph{sigmoid} gate, empowering tokens to activate a variable expert count based on complexity.

\paragraph{Differentiable $\tau$-routing.} 
Let $z = x W_g \in \mathbb{R}^E$ be the unnormalized router logits with learned projection $W_g \in \mathbb{R}^{d_{\text{model}} \times E}$, and $e_{\text{dir}} \in \mathbb{R}^{d_{\text{dir}}}$ be a learnable direction embedding for the translation direction. We compute per-expert probabilities $r$ and a dynamic threshold $\tau$:
\begin{align}
r &= \sigma(z + \epsilon), \quad \epsilon \sim \mathcal{U}(-\eta, \eta), \\
% \tilde{\tau} &= \sigma\bigl([x; e_{\text{dir}}] W_\tau\bigr), \\
\tilde{\tau} &= \sigma\bigl([x; e_{\text{dir}}]^\top W_\tau\bigr), \\
\tau &= \tau_{\min} + (\tau_{\max} - \tau_{\min}) \cdot \tilde{\tau},
\end{align}
where $W_\tau \in \mathbb{R}^{d_{\text{model}} + d_{\text{dir}}}$ projects the concatenated representation, and $\eta$ is a training jitter to prevent routing collapse \cite{fedus2022switch}.
% where $W_\tau \in \mathbb{R}^{(d_{\text{model}} + d_{\text{dir}}) \times 1}$ projects the concatenated representation, and $\eta$ is a training jitter to prevent routing collapse \cite{fedus2022switch}. 
% The bounds $0 < \tau_{\min} < \tau_{\max} < 1$ prevent the router from activating or bypassing all experts.

% A token activates expert $E_i$ if and only if $r_i > \tau$. To backpropagate through this discrete decision, we use a Straight-Through Estimator (STE)~\cite{bengio2013estimating} with steepness $s$:
% \begin{align}
% M^{\text{soft}}_i &= \sigma\bigl(s \cdot (r_i - \tau)\bigr), \\
% M^{\text{hard}}_i &= \mathbb{I}[r_i > \tau], \\
% M_i &= \mathrm{sg}\bigl(M^{\text{hard}}_i - M^{\text{soft}}_i\bigr) + M^{\text{soft}}_i,
% \end{align}
A token activates expert $E_i$ if and only if $r_i > \tau$. To backpropagate through this discrete decision, we use a Straight-Through Estimator (STE)~\cite{bengio2013estimating}:
\begin{align}
M^{\text{soft}}_i &= \sigma\bigl(r_i - \tau\bigr), \\
M^{\text{hard}}_i &= \mathbb{I}[r_i > \tau], \\
M_i &= \mathrm{sg}\bigl(M^{\text{hard}}_i - M^{\text{soft}}_i\bigr) + M^{\text{soft}}_i,
\end{align}
where $\mathrm{sg}(\cdot)$ is the stop-gradient operator. The MoE output is given by:
\begin{equation}
\text{MoE}(x) = \sum_{i=1}^E M_i \, r_i \, E_i(x; D_k).
\end{equation}
During inference, only experts where $M^{\text{hard}}_i = 1$ are activated, rendering per-token compute dynamically data-dependent.

\paragraph{Expert Sparsity Constraints.} 
To explicitly control the computational budget and prevent load imbalance, we enforce a target sparsity $\rho_{\text{target}}$ for active experts. An auxiliary loss aligns the empirical routing sparsity with this constraint to ensure uniform expert utilization:
% \begin{equation}
% \label{eq:sparsity_loss}
% \mathcal{L}_{\text{sp}} = \frac{1}{|\mathcal{B}|} \sum_{x \in \mathcal{B}} \left(\sum_{i=1}^E M_i(x) - \rho_{\text{target}} \right)^2 + \frac{1}{E} \sum_{i=1}^E \left( f_i - \frac{\rho_{\text{target}}}{E} \right)^2,
% \end{equation}

\begin{equation}
\label{eq:sparsity_loss}
\begin{split}
\mathcal{L}_{\text{sp}} = &\frac{1}{|\mathcal{B}|} \sum_{x \in \mathcal{B}} \left(\sum_{i=1}^E M_i(x) - \rho_{\text{target}} \right)^2 \\
&+ \frac{1}{E} \sum_{i=1}^E \left( f_i - \frac{\rho_{\text{target}}}{E} \right)^2.
\end{split}
\end{equation}
where $f_i$ is the activation frequency of expert $i$ within batch $\mathcal{B}$, masking padded positions for accurate sparsity estimation.

\subsection{Dual-Consistency Objective}
\label{sec:method:dual}

A central challenge in elastic MoEs is ensuring aggressively truncated sub-networks remain performant without downstream fine-tuning. We resolve this via a \emph{dual-consistency} objective tying varying capacity views together.

At every training step, we sample two dimensions $D_A, D_B \sim \mathcal{D}$ with replacement. We perform two forward passes by dynamically slicing expert activations to widths $D_A$ and $D_B$, yielding predictive vocabulary distributions $P_A$ and $P_B$. We define the loss using a Symmetrized Kullback-Leibler (KL) divergence (Jeffreys Divergence):
\begin{equation}
\label{eq:dc}
\mathcal{L}_{\text{dc}} = \frac{1}{2} \Big[ \mathrm{KL}\big(\text{sg}(P_A) \| P_B\big) + \mathrm{KL}\big(\text{sg}(P_B) \| P_A\big) \Big],
\end{equation}
where $\text{sg}(\cdot)$ denotes the stop-gradient operator. This formulation unifies two regularizing paradigms. When $D_A \neq D_B$, it functions as \emph{in-place self-distillation}: the larger capacity view provides a richer target to guide the smaller variant, while the smaller core acts as a structural regularizer for the larger network. When $D_A = D_B$, it defaults to consistency regularization similar to \emph{R-Drop} \cite{wu2021r}, minimizing the divergence between stochastic forward passes of the same sub-network. This bidirectional flow anchors the full-capacity model while preserving the translation quality of truncated variants.

\subsection{Direction-Aware Shared Expert}
\label{sec:method:tcse}

Multi-directional and multi-domain NMT models frequently suffer from negative interference. Following recent successes isolating common knowledge in shared experts \cite{rajbhandari2022deepspeed, dai2024deepseekmoe}, we introduce a single \emph{dense} expert activated for every token in parallel with the sparse router. To render this backbone direction-aware without the parameter overhead of dedicated experts, we dynamically modulate its hidden state via Feature-wise Linear Modulation (FiLM) \cite{perez2018film}. We use the discrete translation direction to index learned modulation vectors $\gamma_c, \beta_c \in \mathbb{R}^{M}$. The DASE computes:
\begin{align}
h^{\text{shr}} &= \mathrm{SiLU}(xW^{\text{shr}}_1) \odot (xW^{\text{shr}}_2), \\
\hat{h}^{\text{shr}} &= h^{\text{shr}} \odot (1 + \gamma_c) + \beta_c, \\
E_{\text{shr}}(x; D_k) &= \alpha \cdot \psi(D_k) \cdot \bigl(\hat{h}^{\text{shr}}_{[:D_k]} W^{\text{shr}}_{3, [:D_k, :]} \bigr),
\end{align}
where $0 < \alpha < 1$ is a static damping factor preventing the shared expert from overpowering the sparsely routed experts. The DASE acts as a stable, grammatical backbone, freeing routed experts to specialize heavily across idiosyncratic domains. 

The final output of the elastic MoE block aggregates the dynamic routed paths with the shared direction-conditioned path:
\begin{equation}
\label{eq:block}
y_{\text{MoE}}(x; D_k) = \sum_{i=1}^{E} M_i r_i E_i(x; D_k) + E_{\text{shr}}(x; D_k).
\end{equation}

\subsection{Training Objective and Inference}
\label{sec:method:loss}

The complete optimization objective minimizes the standard label-smoothed cross-entropy loss alongside our auxiliary routing constraints and dual-consistency regularization. Because we sample two distinct capacity dimensions $D_A$ and $D_B$ at each training step, the cross-entropy and routing losses are calculated independently for each width and averaged. The final loss $\mathcal{L}$ is formulated as:
\begin{equation}
\label{eq:total_loss}
\mathcal{L} = \frac{1}{2} \sum_{D \in \{D_A, D_B\}} \left( \mathcal{L}_{\text{ce}}^{(D)} + \lambda_{\text{sp}} \mathcal{L}_{\text{sp}}^{(D)} \right) + \lambda_{\text{dc}}\mathcal{L}_{\text{dc}},
\end{equation}
where $\mathcal{L}_{\text{ce}}^{(D)}$ and $\mathcal{L}_{\text{sp}}^{(D)}$ denote the cross-entropy and target sparsity computed during the forward pass of width $D$. The scalar weights $\lambda_{\text{sp}}$ and $\lambda_{\text{dc}}$ balance the relative impact of the routing and consistency constraints.

During inference, a target memory footprint is dictated simply by specifying an expert width $D_k \in \mathcal{D}$. The corresponding sub-network is extracted seamlessly via deterministic matrix slicing. Simultaneously, the adaptive sigmoid router dynamically calibrates the active expert count per token. By decoupling static memory capacity from the dynamic compute budget, EDWARD-NMT realizes a highly efficient ``train once, deploy anywhere'' paradigm.

\section{Experimental Setup}
\label{sec:exp}

% \subsection{Datasets}
% We evaluate EDWARD-NMT on three benchmark families.

% \paragraph{PhoMT (En$\leftrightarrow$Vi).} A large-scale parallel corpus \cite{doan2021phomt} ($\approx$3M pairs) spanning news, subtitles, legal, and Wikipedia text. We use the official splits with a shared 20k SentencePiece \cite{kudo-richardson-2018-sentencepiece} vocabulary.

% \paragraph{WMT-14 (En$\leftrightarrow$De).} The standard benchmark \cite{bojar2014findings} ($\approx$3.9M filtered pairs from 4.5M original pairs). We use a shared 32k BPE \cite{sennrich2016neural} vocabulary and report on \texttt{newstest2014} for direct comparability.

% \paragraph{Multi-domain De$\leftrightarrow$En.} Following~\citet{zhang2024lightweight}, we use $\approx$1.3M filtered pairs across five domains: \textsc{Koran}, \textsc{IT}, \textsc{Medical}, \textsc{Law}, and \textsc{Subtitles}. Domains are jointly trained using a shared 32k BPE vocabulary, evaluated separately, and macro-averaged.

\subsection{Datasets}
We evaluate EDWARD-NMT on four benchmark families (full dataset and preprocessing details are provided in Appendix~\ref{sec:appendix:data}): (1) \textbf{PhoMT} (En$\leftrightarrow$Vi) \cite{doan2021phomt}, (2) the standard \textbf{WMT-14} (En$\leftrightarrow$De) \cite{bojar2014findings} evaluated on \texttt{newstest2014}, (3) the \textbf{multi-domain} De$\leftrightarrow$En benchmark \cite{zhang2024lightweight} spanning five diverse domains evaluated via macro-averaging, and (4) the multilingual translation \textbf{OPUS-16} dataset following \citet{zhao2024sparse-lingual-smoe}, with 16 languages divided into high ($>0.9$M pairs), low ($<0.1$M pairs) and medium resource.

\subsection{Architecture and Sub-network Family}
\label{sec:exp:arch}
The encoder-decoder backbone has $L_{\text{enc}} = L_{\text{dec}} = 8$ layers, $d_{\text{model}} = 512$, eight attention heads, $E = 4$ routed experts, and one direction-aware shared expert (total of $5$ experts) per MoE layer. Both routed and shared experts have a maximum inner dimension $M = 2048$, governed by the nested family: $\mathcal{D} = \{2048,\,1024,\,512,\,256\}$.
This yields the four variants in Table~\ref{tab:variants}. Total parameters vary across datasets due to vocabulary size. All share \emph{a single set of trained parameters}, differing only by the inference slice $D_k \in \mathcal{D}$.

\begin{table}[!ht]
\centering
\small
\setlength{\tabcolsep}{2pt}
\begin{tabular}{lccc}
\toprule
\textbf{Variant} & $\mathbf{D_k}$ & \textbf{Total Params} & \textbf{Active Params} \\
\midrule
EDWARD-NMT-S  & $256$  & $67-74$M   & $51-59$M \\
EDWARD-NMT-M  & $512$  & $98-105$M  & $66-76$M \\
EDWARD-NMT-L  & $1024$ & $161-168$M & $97-109$M \\
EDWARD-NMT-XL & $2048$ & $287-294$M & $158-175$M \\
\bottomrule
\end{tabular}
\caption{EDWARD-NMT sub-network variants extracted from a single trained model.}
\label{tab:variants}
\end{table}

% \begin{table}[!ht]
% \centering
% \small
% \begin{tabular}{lcc}
% \toprule
% \textbf{Variant} & $\mathbf{D_k}$ & \textbf{Active Params} \\
% \midrule
% EDWARD-NMT-S  & $256$  & $51-59$M \\
% EDWARD-NMT-M  & $512$  & $66-76$M \\
% EDWARD-NMT-L  & $1024$ & $97-109$M \\
% EDWARD-NMT-XL & $2048$ & $158-175$M \\
% \bottomrule
% \end{tabular}
% \caption{EDWARD-NMT sub-network variants extracted from a single trained model.}
% \label{tab:variants}
% \end{table}

\paragraph{Routing \& Consistency.} The direction embedding $e_{\text{dir}} \in \mathbb{R}^{64}$ is learned per translation direction. The router uses jitter $\eta = 0.05$, thresholds $\tau_{\min} = 0.05$ and $\tau_{\max} = 0.60$. The target sparsity is set to $\rho_{\text{target}} = 1.25$. The auxiliary weight is $\lambda_{\text{sp}} = 0.01$, and the DASE damping factor is $\alpha = 0.1$. For dual-consistency ($\lambda_{\text{dc}} = 1$), dimensions $D_A, D_B \sim \mathcal{D}$ are sampled uniformly with replacement per step.

\subsection{Baselines}
We compare against standard dense Transformers (\textsc{Base}/\textsc{Big}) \cite{vaswani2017attention,ott2018scaling}, mBART \cite{liu2020multilingual}, and MoE methods: the Switch Transformer \cite{fedus2022switch} with top-1 token-based routing (\textsc{ST-MoE}) and GShard \cite{lepikhin2020gshard} with top-2 token-based routing (\textsc{GS-MoE}). Both MoE models use \textsc{Transformers Big} backbone for multi-domain benchmark. All models are \emph{trained jointly on all translation directions} to match our setup. For OPUS-16 we report baseline results from~\citet{zhao2024sparse-lingual-smoe}, more MoE baselines are provided in Appendix~\ref{sec:appendix:opus16}.

% \subsection{Implementation Details}
% \label{sec:exp:opt}

% \paragraph{Training.} Models are optimized using AdamW \cite{loshchilov2018decoupled} ($\beta_1{=}0.9$, $\beta_2{=}0.98$), with learning rate warms up linearly from $10^{-6}$ to a peak of $10^{-2}$ over the first $1$k steps. We use an effective batch size of $8{,}192$ sequences (max length $128$). We train for 16k steps on the multi-domain, 35k on OPUS-16, 40k on PhoMT, and 60k on WMT-14, applying a label smoothing factor of $0.1$ and a dropout rate of $0.1$. The complete hyperparameter configuration is detailed in Table~\ref{tab:appendix:hp} (Appendix).

% \paragraph{Inference and Evaluation.} We decode via beam search (width $4$, penalty $0.6$) on TPU v5 pods (Appendix~\ref{sec:appendix}). We report case-sensitive sacreBLEU \cite{post2018call} for PhoMT, WMT-14 and OPUS-16, and tokenized BLEU \cite{papineni-etal-2002-bleu} for multi-domain. All the four EDWARD-NMT variants per row are extracted from the \emph{same} checkpoint to directly compare quality versus parameter count.

\subsection{Implementation Details}
\label{sec:exp:opt}

\paragraph{Training.} All models are trained on a single TPU v5-64 pod using the AdamW optimizer \cite{loshchilov2018decoupled} ($\beta_1{=}0.9$, $\beta_2{=}0.98$). The learning rate warms up linearly from $10^{-6}$ to $10^{-2}$ over the first 1k steps. We use an effective batch size of $8{,}192$ sequences (max length $128$), a dropout rate of $0.1$, and label smoothing of $0.1$. Training spans 16k steps for the multi-domain, 30k for OPUS-16, 40k for PhoMT, and 60k for WMT-14. Complete hyperparameters are listed in Appendix Table~\ref{tab:appendix:hp}.

% \paragraph{Inference and Evaluation.} We decode via beam search (width $4$, length penalty $0.6$). Performance is measured using case-sensitive sacreBLEU \cite{post2018call} for PhoMT, WMT-14, and OPUS-16, and tokenized BLEU \cite{papineni-etal-2002-bleu} for the multi-domain dataset. Crucially, all four EDWARD-NMT variants are extracted and evaluated from the \emph{same} trained checkpoint.
\paragraph{Inference and Evaluation.} We decode via beam search (width $4$, length penalty $0.6$). Performance is measured using case-sensitive sacreBLEU \cite{post2018call} for PhoMT, WMT-14, and OPUS-16, and tokenized BLEU \cite{papineni-etal-2002-bleu} for the multi-domain dataset. We also report COMET \cite{rei-etal-2020-comet} for semantic fidelity on WMT-14 and multi-domain dataset. chrF2 \cite{popovic-2016-chrf} on WMT-14 and TER \cite{snover2006study-ter} on PhoMT results are detailed in Appendix~\ref{sec:appendix:wmt14} and~\ref{sec:appendix:abl-phomt}. All reported improvements over baselines are statistically significant (Appendix~\ref{sec:appendix:eval}). Edge inference latency and throughput are profiled using compiled LiteRT checkpoints (Appendix~\ref{sec:appendix:compute}). Crucially, all four \textsc{EDWARD-NMT} variants are extracted and evaluated from the \emph{same} checkpoint.

\section{Results}
\label{sec:result}

\subsection{Main Translation Quality}
\label{sec:result:main}

\paragraph{PhoMT and WMT-14.}
% Table~\ref{tab:main_phomt_wmt} reports sacreBLEU on PhoMT (En$\leftrightarrow$Vi) and WMT-14 (En$\leftrightarrow$De). The full \textsc{EDWARD-NMT-XL} matches or exceeds \textsc{Transformer Big} across all directions while activating fewer parameters, improving by up to $+1.30$ BLEU on PhoMT and $+1.48$ on WMT-14. Furthermore, it surpasses much larger mBART models ($448-610$M parameters) on both benchmarks. The smaller sub-networks extracted from the same checkpoint remain highly competitive: \textsc{EDWARD-NMT-S} ($51-59$M active params) outperforms the dense \textsc{Transformer Base} by $+0.7$ to $+3.2$ BLEU at a comparable parameter footprint, performing on par with \textsc{Transformer Big} on PhoMT.
% Table~\ref{tab:main_phomt_wmt} reports sacreBLEU on PhoMT (En$\leftrightarrow$Vi) and WMT-14 (En$\leftrightarrow$De). \textsc{EDWARD-NMT-XL} exceeds \textsc{Transformer Big} across all directions while activating fewer parameters, improving by $+1.30$ BLEU on PhoMT (Vi$\rightarrow$En) and $+1.48$ on WMT-14 (De$\rightarrow$En). Furthermore, it surpasses much larger mBART models across all directions on WMT-14 and En$\rightarrow$Vi, while remaining highly competitive on Vi$\rightarrow$En despite using more $3\times$ fewer active parameters. The smaller sub-networks extracted from the same checkpoint remain highly competitive: \textsc{EDWARD-NMT-S} ($51-59$M active params) outperforms the dense \textsc{Transformer Base} by $+0.69$ to $+3.18$ BLEU at a comparable parameter footprint, performing on par with \textsc{Transformer Big} on PhoMT.

Table~\ref{tab:main_phomt_wmt} reports sacreBLEU on PhoMT (En$\leftrightarrow$Vi) and WMT-14 (En$\leftrightarrow$De). \textsc{EDWARD-NMT-XL} exceeds \textsc{Transformer Big} across all directions while activating fewer parameters, improving by $+1.30$ BLEU on PhoMT (Vi$\rightarrow$En) and $+1.48$ on WMT-14 (De$\rightarrow$En). Furthermore, it surpasses much larger mBART models across all directions on En$\rightarrow$Vi, while remaining highly competitive on Vi$\rightarrow$En despite using $3\times$ fewer active parameters. The smaller sub-networks extracted from the same checkpoint maintain robust performance: \textsc{EDWARD-NMT-S} ($51 - 59$M active parameters) outperforms the dense \textsc{Transformer Base} by $+0.69$ to $+3.18$ BLEU at a comparable parameter footprint, performing on par with \textsc{Transformer~Big} on PhoMT. 

% \begin{table*}[t]
% \centering
% \small
% \setlength{\tabcolsep}{4pt}
% \begin{tabular}{lcccc cccc}
% \toprule
% & \multicolumn{4}{c}{\textbf{PhoMT}} & \multicolumn{4}{c}{\textbf{WMT-14}} \\
% \cmidrule(lr){2-5}\cmidrule(lr){6-9}
% Method & Params & Active & En--Vi & Vi--En & Params & Active & En--De & De--En \\
% \midrule
% \textsc{Transformer Base} & 65M  & 65M  & 42.12 & 37.19 & 61M  & 61M  & 25.65 & 30.39 \\
% \textsc{Transformer Big} & 213M & 213M & 42.94 & 37.83 & 210M & 210M & \underline{28.5} & \underline{33.16} \\
% mBART            & 448M & 448M & 43.46 & \textbf{39.78} & --   & --   & --    & --    \\
% \midrule
% EDWARD-NMT-S  & 67M  & 54M  & 42.91 & 37.88 & 74M  & 61M  & 27.39 & 32.41 \\
% EDWARD-NMT-M  & 98M  & 73M  & 43.35 & 38.50 & 105M & 80M  & 28.05 & 32.92 \\
% EDWARD-NMT-L  & 161M & 111M & \underline{43.61} & 38.93 & 168M & 118M & \underline{28.5} & 33.13 \\
% EDWARD-NMT-XL & 287M & 187M & \textbf{43.79} & \underline{39.13} & 294M & 193M & \textbf{28.62} & \textbf{33.44} \\
% \bottomrule
% \end{tabular}
% \caption{Main results on PhoMT (En$\leftrightarrow$Vi, sacreBLEU) and WMT-14 (En$\leftrightarrow$De, tokenized BLEU). All four EDWARD-NMT variants are extracted from a \emph{single} trained checkpoint. ``Active'' reports the average activated parameters with the proposed sigmoid router. The \textbf{best} and \underline{second best} scores are highlighted in bold and underlined.}
% \label{tab:main_phomt_wmt}
% \end{table*}

\begin{table*}[!ht]
\centering
\small
\begin{tabular}{lcccc cccc}
\toprule
& \multicolumn{4}{c}{\textbf{PhoMT}} & \multicolumn{4}{c}{\textbf{WMT-14}} \\
\cmidrule(lr){2-5}\cmidrule(lr){6-9}
Method & Params & Active & En$\rightarrow$Vi & Vi$\rightarrow$En & Params & Active & En$\rightarrow$De & De$\rightarrow$En \\
\midrule
\textsc{Transformer Base} & 65M  & 65M  & 42.12 & 37.19 & 61M  & 61M  & 24.56 & 28.76 \\
\textsc{Transformer Big} & 213M & 213M & 42.94 & 37.83 & 210M & 210M & 27.53 & 31.48 \\
mBART$^{\dagger}$        & 448M & 448M & 43.46 & \textbf{39.78} & 610M & 610M & 24.90  & 32.51 \\
\midrule
EDWARD-NMT-S  & 67M  & 51M  & 42.91 & 37.88 & 74M  & 59M  & 26.85 & 31.94 \\
EDWARD-NMT-M  & 98M  & 66M  & 43.35 & 38.50 & 105M & 76M  & 27.49 & 32.45 \\
EDWARD-NMT-L  & 161M & 97M & \underline{43.61} & 38.93 & 168M & 109M & \underline{27.96} & \underline{32.65} \\
EDWARD-NMT-XL & 287M & 158M & \textbf{43.79} & \underline{39.13} & 294M & 175M & \textbf{28.03} & \textbf{32.96} \\
\bottomrule
\end{tabular}
\caption{Main results on PhoMT (En$\leftrightarrow$Vi) and WMT-14 (En$\leftrightarrow$De) measured in sacreBLEU. ``Active'' denotes average parameters activated by our sigmoid router. Best and second-best scores are \textbf{bolded} and \underline{underlined}, respectively. $^{\dagger}$For mBART on WMT-14, we use a pretrained many-to-many model (\texttt{mbart-large-50-many-to-many-mmt}).}
\label{tab:main_phomt_wmt}
\end{table*}

% \paragraph{Multi-domain De$\leftrightarrow$En.}
% \input{Chapter/tables/main_multidomain_avg}
% Table~\ref{tab:main_multi_avg} compares EDWARD-NMT to standard dense and MoE baselines on the multi-domain benchmark, reporting tokenized BLEU averaged across both directions. \textsc{EDWARD-NMT-XL} achieves the highest average BLEU (40.75), consistently outperforming the strong \textsc{Transformer Big} baseline (40.13) across all five domains while requiring significantly fewer active parameters (161M vs. 210M). This efficiency extends to smaller sub-networks: \textsc{EDWARD-NMT-L} surpasses \textsc{Transformer Big} using less than half the active parameters (102M). Additionally, our smallest variant, \textsc{EDWARD-NMT-S} (57M active params), provides a substantial +0.49 BLEU improvement over \textsc{Transformer Base} (38.16) and achieves comparable performance to the much heavier \textsc{GS-MoE} (260M active params). Crucially, EDWARD-NMT supplies all four sizes from a \emph{single} training run, whereas baselines require separately trained models.

\paragraph{Multi-domain De$\leftrightarrow$En.}

\begin{table*}[t]
\centering
\small
\begin{tabular}{lcc cccccc cc}
\toprule
& & & \multicolumn{6}{c}{\textbf{Multi-domain}} & \multicolumn{2}{c}{\textbf{OPUS-16}} \\
\cmidrule(lr){4-9} \cmidrule(lr){10-11}
Method & Params & Active & \textsc{Kor} & \textsc{IT} & \textsc{Med} & \textsc{Law} & \textsc{Sub} & \textsc{Avg} & En$\rightarrow$XX & XX$\rightarrow$En \\
\midrule
\textsc{Transformer Base} & 61M & 61M & 20.53 & 40.01 & 51.25 & 53.72 & 25.29 & 38.16 & 26.92$^\dagger$ & 30.67$^\dagger$ \\
\textsc{Transformer Big}  & 210M & 210M & 22.55 & \underline{41.84} & \underline{53.35} & 56.22 & 26.73 & 40.13 & - & - \\
\textsc{ST-MoE}           & 360M & 210M & 21.73 & 41.00 & 52.01 & 55.69 & 26.18 & 39.32 & 29.68$^\dagger$ & 33.89$^\dagger$ \\
\textsc{GS-MoE}           & 360M & 260M & 20.70 & 41.35 & 52.31 & 54.56 & 24.72 & 38.72 & 28.28$^\dagger$ & 32.31$^\dagger$ \\
\midrule
\textsc{EDWARD-NMT-S}   &  74M & 57M  & 21.61 & 40.07 & 51.01 & 53.99 & 26.56 & 38.65 & 29.55 & 32.99 \\
\textsc{EDWARD-NMT-M}   & 105M & 72M  & 22.72 & 40.87 & 52.11 & 55.21 & 26.88 & 39.56 & 30.73 & 34.84 \\
\textsc{EDWARD-NMT-L}   & 168M & 102M & \underline{23.56} & 41.53 & 52.95 & \underline{56.27} & \underline{27.13} & \underline{40.28} & \underline{31.64} & \underline{35.82} \\
\textsc{EDWARD-NMT-XL}  & 294M & 161M & \textbf{24.06} & \textbf{41.90} & \textbf{53.45} & \textbf{57.05}  & \textbf{27.30} & \textbf{40.75} & \textbf{32.27} & \textbf{36.30} \\
\bottomrule
\end{tabular}
\caption{Experimental results on the multi-domain (De$\leftrightarrow$En) averaged across both directions and multilingual OPUS-16 benchmarks (English to any En$\rightarrow$XX and any to English XX$\rightarrow$En). The ``Params'' and ``Active'' columns report footprints for the multi-domain setup. $^\dagger$Reported from~\citet{zhao2024sparse-lingual-smoe}. Extended OPUS-16 comparisons are provided in Appendix~\ref{sec:appendix:opus16}.}
\label{tab:main_multi_avg}
\end{table*}
% Table~\ref{tab:main_multi_avg} compares EDWARD-NMT to dense and MoE baselines on the multi-domain benchmark. Our \textsc{EDWARD-NMT-XL} achieves the highest average BLEU (40.75), outperforming \textsc{Transformer Big} (40.13) across all five domains with significantly fewer active parameters (161M vs.\ 210M). This efficiency extends to smaller sub-networks: \textsc{EDWARD-NMT-L} surpasses \textsc{Transformer Big} using under half the active parameters (102M), while the smallest \textsc{EDWARD-NMT-S} (57M) improves upon \textsc{Transformer Base} by $+0.49$ BLEU and rivals the much heavier \textsc{GS-MoE} (260M). Crucially, EDWARD-NMT extracts all four sizes from a \emph{single} training run, whereas baselines require separate training.
Table~\ref{tab:main_multi_avg} compares EDWARD-NMT to dense and MoE baselines on the multi-domain benchmark. Our \textsc{EDWARD-NMT-XL} achieves the highest average BLEU, outperforming \textsc{Transformer Big} by $+0.62$ BLEU across all five domains with significantly fewer active parameters (161M vs.\ 210M). This efficiency extends to smaller sub-networks: \textsc{EDWARD-NMT-L} surpasses \textsc{Transformer Big} using under half the active parameters (102M), while the smallest \textsc{EDWARD-NMT-S} (57M) improves upon \textsc{Transformer Base} by $+0.49$ BLEU and rivals the much heavier \textsc{GS-MoE} (260M). 

\paragraph{Multilingual OPUS-16.}
% Demonstrating robust scalability, our architecture also excels on the massively multilingual OPUS-16 dataset (Table~\ref{tab:main_multi_avg}). Even the most compressed sub-network, \textsc{EDWARD-NMT-S}, consistently outperforms \textsc{ST-MoE} and \textsc{GS-MoE} in both translation directions. The full-capacity \textsc{EDWARD-NMT-XL} further extends this lead, peaking at $32.27$ and $36.30$ BLEU. An extended comparison against a wider range of baselines is provided in Appendix~\ref{sec:appendix:opus16}.
Demonstrating robust scalability, our architecture also excels on the massively multilingual OPUS-16 dataset (Table~\ref{tab:main_multi_avg}). While our most compressed sub-network, \textsc{EDWARD-NMT-S}, remains competitive, scaling to just the \textsc{EDWARD-NMT-M} variant is sufficient to consistently outperform both \textsc{ST-MoE} and \textsc{GS-MoE} across both translation directions. The full-capacity \textsc{EDWARD-NMT-XL} further extends this lead. An extended comparison against a wider range of baselines is provided in Appendix~\ref{sec:appendix:opus16}.

\paragraph{Semantic Quality}

\begin{table}[t]
\centering
\small
\setlength{\tabcolsep}{4pt}
\begin{tabular}{l cc cc}
\toprule
& \multicolumn{2}{c}{\textbf{WMT-14}} & \multicolumn{2}{c}{\textbf{Multi-domain}} \\
\cmidrule(lr){2-3} \cmidrule(lr){4-5}
Method & Active & \textsc{Avg} & Active & \textsc{Avg} \\
\midrule
\textsc{Transformer Base} & 61M & 80.78 & 61M & 80.20 \\
\textsc{Transformer Big}  & 210M & 83.69 & 210M & \underline{81.62} \\
mBART$^{\dagger}$         & 610M & 83.56 & - & - \\
\textsc{ST-MoE}           & - & - & 210M & 80.95 \\
\textsc{GS-MoE}           & - & - & 260M & 78.20 \\
\midrule
\textsc{EDWARD-NMT-S}  & 59M & 83.48 & 57M & 80.96 \\
\textsc{EDWARD-NMT-M}  & 76M & 83.98 & 72M & 81.32 \\
\textsc{EDWARD-NMT-L}  & 109M & \underline{84.12} & 102M & \underline{81.62} \\
\textsc{EDWARD-NMT-XL} & 175M & \textbf{84.30} & 161M & \textbf{81.80} \\
\bottomrule
\end{tabular}
\caption{Semantic-level translation quality measured in COMET-22. Scores are averaged across all translation directions (and domains for the Multi-domain benchmark). Best and second-best scores are \textbf{bolded} and \underline{underlined}. $^{\dagger}$mBART is a pretrained many-to-many model.}
\label{tab:comet_avg}
\end{table}
Beyond surface-level $n$-gram matching, we evaluate semantic integrity using COMET-22\footnote{\url{https://huggingface.co/Unbabel/wmt22-comet-da}} \cite{rei2022comet}. Table~\ref{tab:comet_avg} presents the average scores across all evaluation directions and domains. On WMT-14, our highly compressed \textsc{EDWARD-NMT-S} (59M active parameters) significantly outperforms \textsc{Transformer Base} (61M active) by an average margin of $+2.70$ COMET. Additionally, our \textsc{EDWARD-NMT-M} (76M) and \textsc{EDWARD-NMT-L} (109M) variants consistently exceed the 210M-parameter \textsc{Transformer Big}. At full capacity, \textsc{EDWARD-NMT-XL} establishes a new Pareto frontier by surpassing \textsc{Transformer Big} by $+0.61$ COMET on average. This semantic robustness extends to the Multi-domain benchmark, where heavily parameterized models such as \textsc{GS-MoE} (260M) experience severe semantic degradation (78.20 COMET). In stark contrast, our \textsc{EDWARD-NMT-S} model (57M) maintains a robust score of 80.96 COMET. This performance is comparable to the 210M-parameter \textsc{ST-MoE} and outperforms \textsc{GS-MoE} by $+2.76$ points. Furthermore, the \textsc{EDWARD-NMT-XL} model achieves the highest overall score (81.80), demonstrating that our architecture strictly preserves meaning even within aggressively truncated sub-networks.

% \paragraph{Quality vs. Active Parameters.}
% \input{Chapter/tables/main_exp_density}
% Driven by our adaptive sigmoid routing, which activates only 1.34 to 1.70 experts per token (Table~\ref{tab:avg_active_experts}), \textsc{EDWARD-NMT} establishes an efficient Pareto frontier. The extracted sub-networks consistently outperform dense baselines like \textsc{Transformer Big} and mBART while using up to $3.6\times$ fewer active parameters. For instance, \textsc{EDWARD-NMT-L} surpasses \textsc{Transformer Big} utilizing less than half the active capacity (102M vs. 210M) and higher generation throughput (61.5 vs. 48.0 TPS). This structural compactness translates effectively to practical on-device acceleration. While experiencing only a marginal drop in translation quality, the compact \textsc{S} variant reduces the total parameter footprint by $4\times$ and delivers a $2.2\times$ speedup over the full XL variant (Appendix Table~\ref{tab:inference_perf}), with all sizes extracted from a single checkpoint.

\paragraph{Quality vs. Active Parameters.}
\begin{table}[t]
\centering
\small
\setlength{\tabcolsep}{2.5pt}
\begin{tabular}{lcccc}
\toprule
Method & \textbf{PhoMT} & \textbf{WMT-14} & \textbf{MD} & \textbf{OPUS-16} \\
\midrule
EDWARD-NMT-S  & 1.44 & 1.64 & 1.37 & 1.34 \\
EDWARD-NMT-M  & 1.45 & 1.66 & 1.37 & 1.35 \\
EDWARD-NMT-L  & 1.47 & 1.69 & 1.38 & 1.37\\
EDWARD-NMT-XL & 1.51 & 1.70 & 1.39 & 1.39 \\
\bottomrule
\end{tabular}
% \caption{Average number of active routed experts per token across PhoMT, WMT-14, and MD (Multi-domain) benchmark.}
\caption{Average active \textbf{routed experts} per token across the PhoMT, WMT-14, Multi-domain (MD) and OPUS-16 benchmarks (out of $E=4$ routed experts, excluding the fixed shared expert).}
\label{tab:avg_active_experts}
\end{table}
Driven by our adaptive sigmoid routing, which activates only 1.34 to 1.70 experts per token (Table~\ref{tab:avg_active_experts}), \textsc{EDWARD-NMT} establishes an efficient Pareto frontier. Crucially, this number of active experts is explicitly controlled by the target density $\rho$ (Table~\ref{tab:density_settings}). The extracted sub-networks consistently outperform dense baselines like \textsc{Transformer Big} and mBART while using up to $3.6\times$ fewer active parameters. For instance, \textsc{EDWARD-NMT-L} surpasses \textsc{Transformer Big} utilizing less than half the active capacity (102M vs. 210M) and higher generation throughput (61.5 vs. 48.0 TPS). This structural compactness translates effectively to practical on-device acceleration. While experiencing only a marginal drop in translation quality, the compact \textsc{S} variant reduces the total parameter footprint by $4\times$ and delivers a $2.2\times$ speedup over the full XL variant (Appendix Table~\ref{tab:inference_perf}), with all sizes extracted from a single checkpoint.

\subsection{Ablation Studies}
\label{sec:result:abl}

All ablations are run on the multi-domain benchmark (averaged across En$\leftrightarrow$De) and report the tokenized BLEU of every sub-network (XL/L/M/S) extracted from the same training run. PhoMT ablations are provided in Appendix~\ref{sec:appendix:abl-phomt}.

\paragraph{Contributions of Each Component.}
Table~\ref{tab:abl_avg} incrementally adds EDWARD-NMT components to a Top-2 MoE baseline. Adding Matryoshka Experts (ME) yields immediate gains, peaking the XL capacity at 40.96. While fully integrating the Sigmoid router and Dual-Consistency (DC) causes a slight XL quality drop on this benchmark, this represents a deliberate, highly favorable trade-off. The adaptive router drastically reduces the active compute budget (Top-2 $\rightarrow$ $\sim$1.38 experts), while DC prevents smaller sub-networks from collapsing. Consequently, the full system (\textbf{+ ME + DC + Sigmoid}) provides the optimal configuration, maintaining highly competitive XL performance at significantly lower inference costs, while maximizing translation quality for the L, M, and S variants. Ablations on PhoMT (Appendix Table~\ref{tab:abl_contrib_avg}) confirm the efficacy that it improves all capacities performance.

% \begin{table}[t]
% \centering
% \small
% \setlength{\tabcolsep}{5pt}
% \begin{tabular}{lcccc}
% \toprule
% & \multicolumn{4}{c}{\textbf{Avg (En$\leftrightarrow$De)}} \\
% \cmidrule(lr){2-5}
% Setting & XL & L & M & S \\
% \midrule
% MoE ($k{=}2$)    & 40.61 & 40.09 & \textbf{39.77} & \textbf{38.70} \\
% + ME & \textbf{40.96} & \underline{40.26} & 39.35 & 38.34 \\
% + ME + Sigmoid & 40.72 & 40.23 & 39.20 & 38.16 \\
% + ME + DC  & 40.66 & 40.16 & 39.36 & 38.49 \\
% \textbf{+ ME + DC + Sigmoid} & \underline{40.75} & \textbf{40.28} & \underline{39.56} & \underline{38.65} \\
% \bottomrule
% \end{tabular}
% \caption{Ablation results on the Multi-domain benchmark, averaged over both translation directions. The \textbf{best} and \underline{second best} scores are highlighted in bold and underlined.}
% \label{tab:abl_avg}
% \end{table}

\begin{table}[t]
\centering
\small
\setlength{\tabcolsep}{5pt}
\begin{tabular}{lcccc}
\toprule
& \multicolumn{4}{c}{\textbf{En$\leftrightarrow$De}} \\
\cmidrule(lr){2-5}
Setting & XL & L & M & S \\
\midrule
MoE ($k{=}2$)    & 40.35 & 39.64 & 39.21 & 38.24 \\
+ ME & \textbf{40.96} & \underline{40.26} & 39.35 & 38.34 \\
+ ME + Sigmoid & 40.72 & 40.23 & 39.20 & 38.16 \\
+ ME + DC  & 40.66 & 40.16 & \underline{39.36} & \underline{38.49} \\
\textbf{+ ME + DC + Sigmoid} & \underline{40.75} & \textbf{40.28} & \textbf{39.56} & \textbf{38.65} \\
\bottomrule
\end{tabular}
\caption{Ablation results on the Multi-domain benchmark, averaged over both translation directions. \textit{All settings use shared expert (DASE)}.}
\label{tab:abl_avg}
\end{table}

\begin{table}[!ht]
\centering
\small
\begin{tabular}{lcccc}
\toprule
& \multicolumn{4}{c}{\textbf{En$\leftrightarrow$De}} \\
\cmidrule(lr){2-5}Setting & XL & L & M & S \\
\midrule
No SE     & 39.75 & 39.40 & 38.84 & 38.13 \\
Common SE & \underline{39.95} & \underline{39.73} & \underline{39.12} & \underline{38.46} \\
\textbf{DASE (ours)}          & \textbf{40.74} & \textbf{40.26} & \textbf{39.39} & \textbf{38.72} \\
\bottomrule
\end{tabular}
\caption{Ablation on the shared expert (SE) routing strategies on the Multi-domain benchmark, averaged over both translation directions.}
\label{tab:abl_shared_expert_avg}
\end{table}

\begin{table}[!ht]
\centering
\small
\begin{tabular}{lcccc}
\toprule
& \multicolumn{4}{c}{\textbf{En$\leftrightarrow$De}} \\
\cmidrule(lr){2-5}
Setting & XL & L & M & S \\
\midrule
No loss & 40.72 & \underline{40.23} & 39.20 & 38.16 \\
R-Drop & 40.00 & 39.73 & 39.19 & 38.47 \\
Sym.\ JSD & \underline{40.80} & 40.17 & 39.19 & 38.19 \\
Asym.\ KL & 40.76 & 40.14 & \underline{39.50} & \underline{38.59} \\
Asym.\ KL (Max) & \textbf{41.04} & 39.89 & 39.14 & 38.39 \\
\textbf{DC Loss (ours)} & 40.75 & \textbf{40.28} & \textbf{39.56} & \textbf{38.65} \\
\bottomrule
\end{tabular}
\caption{Ablation on the Dual-Consistency loss (DC Loss) on the Multi-domain benchmark, averaged over both translation directions.}
\label{tab:abl_kl_avg}
\end{table}

\paragraph{Impact of Shared Expert Routing.}
Table~\ref{tab:abl_shared_expert_avg} isolates the effect of the shared expert strategy. Compared to a standard MoE without shared experts (No SE), introducing a Common Shared Expert (Common SE) provides modest gains ($+0.2$ to $+0.3$ BLEU). However, our Direction-Aware Shared Expert (DASE) unlocks substantial improvements across the board, yielding approximately $+1.0$ BLEU on XL and $+0.6$ BLEU on S over the No SE baseline. This confirms that intelligently routing common knowledge is vital for maximizing multi-domain translation quality.

% \paragraph{Choice of Consistency Loss.}
% Our proposed Dual-Consistency (\textbf{DC Loss}) introduces a bidirectional alignment mechanism critical for elastic architectures. As shown in Table~\ref{tab:abl_kl_avg}, standard R-Drop isolates identical widths, which limits cross-scale knowledge transfer and degrades the flagship capacity ($-0.6$ BLEU on L). While unidirectional distillation (Asym.\ KL) mitigates this by treating the larger sampled width as a detached teacher, it lacks mutual regularization. Furthermore, anchoring strictly to the full-capacity model (Asym.\ KL Max) yields sub-optimal intermediate variants. By coupling capacity views - with larger variants serving as distillation targets and smaller cores acting as regularizers - our DC Loss establishes a balance that maximizes quality across all sub-networks.

\paragraph{Choice of Consistency Loss.}
Our proposed Dual-Consistency (\textbf{DC Loss}) introduces a bidirectional alignment mechanism critical for elastic architectures. As shown in Table~\ref{tab:abl_kl_avg}, standard R-Drop isolates identical widths, which limits cross-scale knowledge transfer and degrades the flagship capacity ($-0.6$ BLEU on L). A symmetric Jensen-Shannon Divergence (Sym.\ JSD)~\cite{jsd} is similarly ineffective, providing negligible gains for smaller sub-networks. While unidirectional distillation (Asym.\ KL) mitigates this by treating the larger sampled width as a detached teacher, it lacks mutual regularization. Furthermore, anchoring strictly to the full-capacity model (Asym.\ KL Max) yields sub-optimal intermediate variants. By coupling capacity views - with larger variants serving as distillation targets and smaller cores acting as regularizers - our DC Loss establishes a balance that maximizes quality across all sub-networks.

\paragraph{Elasticity of Model Capacities.}
EDWARD-NMT's structural elasticity enables fluid extraction of sub-networks at arbitrary granularities. As detailed in Table~\ref{tab:main_all_datasets} (Appendix), despite being optimized on only four discrete views ($\mathcal{D} = \{2048, 1024, 512, 256\}$), the capacity space seamlessly interpolates. Unseen intermediate sub-networks (e.g., $D=1536$ or $768$) maintain robust translation quality and strictly follow a monotonic compute-performance trade-off without post-hoc fine-tuning. However, this elasticity is bounded by the trained envelope; extrapolating below the minimum trained threshold ($D=128$) triggers a severe performance breakdown, confirming that structural integrity requires optimized geometric bounds.

\paragraph{Effect of Variance Preservation.} 
Removing the deterministic rescaling factor $\psi(D_k)$ consistently degrades translation quality across all extracted sub-networks (Table~\ref{tab:abl_variance_preservation}, Appendix). This performance drop becomes significantly more pronounced at smaller capacities ($-0.37$ BLEU for $D=256$), confirming that explicitly stabilizing down-projected activation magnitudes is crucial for the robust joint training of nested structures.

\section{Conclusion}
\label{sec:conclusion}

% We introduced \textbf{EDWARD-NMT}, an elastic MoE architecture realizing a highly efficient \emph{train-once, deploy-anywhere} paradigm for NMT. By synergizing Matryoshka-style elastic experts, an adaptive sigmoid router, and a direction-aware shared expert, our framework decouples the static memory footprint from the dynamic computational budget. Crucially, a symmetric dual-consistency objective aligns predictive distributions across nested widths, ensuring aggressively truncated sub-networks retain high translation quality without post-hoc fine-tuning. 
We introduced \textbf{EDWARD-NMT}, an elastic MoE architecture realizing an efficient \emph{train-once, deploy-anywhere} paradigm for NMT. By synergizing Matryoshka-style elastic experts, an adaptive sigmoid router, and a direction-aware shared expert, our framework decouples the static memory footprint from the dynamic computational budget. Crucially, a symmetric dual-consistency objective aligns predictive distributions across nested widths, ensuring truncated sub-networks retain high quality without post-hoc fine-tuning. 
Extensive evaluations across PhoMT, WMT-14, multilingual OPUS-16, and a multi-domain benchmark validate our approach. The full-capacity EDWARD-NMT matches or exceeds strong MoE baselines, while the seamlessly extracted smaller sub-networks remain highly competitive, outperforming comparably sized dense Transformers. Overall, coupling structural elasticity with dynamic routing offers a practical, scalable path to resource-aware sequence modeling and adaptable on-device deployment.

\section*{Limitations}
\label{sec:limitations}

Despite its effectiveness, EDWARD-NMT has several limitations that offer avenues for future research:

\paragraph{Architectural Scope.} We evaluate EDWARD-NMT strictly on encoder-decoder architectures to demonstrate its efficacy for machine translation targeted at resource-constrained edge devices. Extending this elastic formulation to massively multilingual settings, decoder-only large language models, or broader sequence generation tasks remains an important direction for future work.

\paragraph{Dimensions of Elasticity.} Our design prioritizes structural flexibility within the feed-forward networks, as they constitute the vast majority of parameters in MoE models. Although EDWARD-NMT achieves robust elasticity for these capacities within the optimized training envelope, we have not yet explored elasticity along orthogonal architectural axes. Future work will investigate integrating early exiting (dynamic depth) and variable hidden state dimensions ($d_{\text{model}}$) to realize a fully elastic, multi-dimensional architecture capable of simultaneous depth-and-width adaptation.

% \paragraph{Training Computational Overhead.} While our framework achieves extreme efficiency at inference, this comes at the cost of increased computational overhead during training. Specifically, our symmetric dual-consistency objective necessitates two distinct forward passes per optimization step to align the nested capacity views. Future work could explore more efficient, single-pass techniques or sparse gradient approximations to reduce the wall-clock training time.

% \paragraph{Bounds of Extrapolation.} EDWARD-NMT demonstrates robust interpolation capabilities within the explicitly optimized dimensional envelope ($\mathcal{D}$). However, as observed in our ablation studies, the model's structural integrity degrades sharply when extrapolated below the minimum trained capacity. Investigating continuous width formulations or dynamically expanding the lower bounds of the Matryoshka dimensions without destabilizing the shared backbone remains a challenge for extreme-compression scenarios.

\paragraph{Extreme Compression.} Although EDWARD-NMT interpolates robustly within its trained envelope ($\mathcal{D}$), performance inherently degrades when extrapolating below the minimum trained capacity. Investigating continuous width formulations or dynamically expanding these lower bounds without destabilizing the shared backbone remains a key direction for extreme-compression scenarios.

\section*{Ethical Considerations}
This work uses only publicly available NMT benchmarks with no personally identifiable or sensitive data, and all baselines are cited and reproduced under their respective licenses. Like any NMT system, EDWARD-NMT may propagate biases or inaccuracies from training corpora and should not inform high-stakes decisions without human review. We foresee no direct harmful applications; by improving parameter and compute efficiency, our work lowers the barrier to deploying translation systems on resource-constrained devices, improving accessibility to translation technology.

\section*{Acknowledgments}
This research was supported with Cloud TPUs from Google's TPU Research Cloud (TRC) program. We would like to thank the TRC team for providing the computational resources used in this work.


\bibliography{custom}

\appendix

\section{Experimental Details}
\label{sec:appendix}

% This appendix follows the \emph{Responsible NLP Research} checklist and provides every detail needed to reproduce the numbers reported in Section~\ref{sec:result} from publicly available data and the formulas of Section~\ref{sec:method}. We do not plan to release source code, checkpoints, or generated outputs. The remainder of this appendix is therefore organised so that an independent re-implementation can be carried out solely from the description below.

% This appendix provides the comprehensive experimental details required to reproduce the results reported in Section~\ref{sec:result}. We outline the datasets, training configurations, and hyperparameters necessary to facilitate an independent re-implementation of the methodology described in Section~\ref{sec:method} using publicly available data.

This appendix provides the comprehensive experimental details required to reproduce the results reported in Section~\ref{sec:result}. We outline the datasets, training configurations, and hyperparameters necessary to independently re-implement the methodology~(Section~\ref{sec:method}) using publicly available data.

\subsection{Datasets and Preprocessing}
\label{sec:appendix:data}

The datasets are summarized in Table~\ref{tab:appendix:data}.

\begin{table}[!ht]
\centering
\small
\setlength{\tabcolsep}{3.5pt}
\begin{tabular}{lccccc}
\toprule
\textbf{Dataset} & \textbf{Lang} & \textbf{Train} & \textbf{Dev} & \textbf{Test} & \textbf{Vocab} \\
\midrule
PhoMT & En$\leftrightarrow$Vi          & 2{,}977k & 19{,}151 & 19{,}151 & 20k SPM \\
WMT-14 & En$\leftrightarrow$De         & 3{,}906k & 3{,}000  & 3{,}003  & 32k BPE \\
MD & De$\leftrightarrow$En   & 1{,}284k & 10{,}000 & 10{,}000 & 32k BPE \\
OPUS-16 & 16 lang   & 17{,}559k & 30{,}000 & 30{,}000 & 32k SPM \\
\bottomrule
\end{tabular}
\caption{Dataset sizes (sentence pairs) and shared vocabularies. \textsc{SPM}: SentencePiece. The ``Train'' column reports the final number of filtered sentence pairs used for model training. The multi-domain (MD) training set is the concatenation of five diverse domains (Koran, IT, Medical, Law, Subtitles) following \citet{zhang2024lightweight}. The OPUS-16 dataset includes 16 languages following~\citet{zhao2024sparse-lingual-smoe}.}
\label{tab:appendix:data}
\end{table}

\paragraph{PhoMT (En$\leftrightarrow$Vi).}
A large-scale parallel corpus \cite{doan2021phomt} comprising approximately 3M sentence pairs across domains such as news, subtitles, legal documents, and Wikipedia. We use the full PhoMT corpus, utilizing the official training pairs directly. For subword tokenization, we train a shared SentencePiece \cite{kudo-richardson-2018-sentencepiece} model with a vocabulary size of $20,000$ tokens strictly on the training set, which is subsequently applied to encode all data splits.

\paragraph{WMT-14 (En$\leftrightarrow$De).}
We utilize the standard WMT-14 training corpora (Europarl v7, Common Crawl, and News Commentary v9), comprising approximately 4.5M original sentence pairs. First, we normalize punctuation and tokenize the text using the Moses toolkit \cite{koehn-etal-2007-moses}\footnote{\url{https://github.com/moses-smt/mosesdecoder}}. Next, we learn a joint Byte-Pair Encoding (BPE) \cite{sennrich2016neural} model with $32,000$ merge operations on the concatenated English and German training data, which is subsequently applied to all data splits. Finally, we filter the training and validation sets to remove sentences longer than $250$ tokens or with a source-target length ratio exceeding $1.5$. This results in roughly 3.9M filtered training pairs. Final translation performance is evaluated on the standard \texttt{newstest2014} test set. The models are trained using a joined dictionary capped at $32,768$ tokens. Our preprocessing pipeline is based on open-source Fairseq\footnote{\url{https://github.com/pytorch/fairseq/tree/master/examples/translation}} \cite{ott2019fairseq}.

\paragraph{Multi-domain De$\leftrightarrow$En.}
Following \citet{zhang2024lightweight}, we evaluate on a multi-domain corpus encompassing five distinct domains: \textsc{Koran}, \textsc{IT}, \textsc{Medical}, \textsc{Law}, and \textsc{Subtitles}. The total filtered size is approximately 1.3M sentence pairs. The tokenization and normalization process follows the exact same Moses-based pipeline as WMT-14. We aggregate the training sets from all five domains across both languages to learn a single, unified BPE model with $32,000$ merge operations. After applying the identical length and ratio filtering (maximum length of $250$, length ratio of $1.5$), the filtered data from all domains is concatenated into a single combined training set and a combined validation set. The models are trained using a joined dictionary capped at $32,768$ tokens, while evaluation is conducted separately on the test sets (each has $2000$ samples) of each domain.

\paragraph{Multilingual OPUS-16.}
The OPUS-16 benchmark \cite{zhao2024sparse-lingual-smoe} comprises 16 diverse languages. We prepare the training, validation, and test data following \citet{zhao2024sparse-lingual-smoe}. The training and validation data are filtered to remove sentences exceeding $100$ tokens. We train a joint SentencePiece unigram model with a base vocabulary of $32,000$ tokens on the training set and subsequently apply it to encode all data splits.

\subsection{Architecture}
\label{sec:appendix:arch}
% The encoder-decoder backbone has $L_{\text{enc}}{=}L_{\text{dec}}{=}8$ layers, $d_{\text{model}}{=}512$, $h{=}8$ attention heads ($d_h{=}64$), pre-LayerNorm with RMSNorm \cite{zhang2019root}, and SwiGLU \cite{shazeer2020glu} non-linearities. RoPE \cite{10.1016/j.neucom.2023.127063-rope} position embeddings are used. MoE blocks replace \emph{every} feed-forward layer in both the encoder and decoder. Each MoE block contains $E{=}4$ routed experts with a maximum inner dimension $M{=}2048$ and the nested family $\mathcal{D}=\{2048,1024,512,256\}$, alongside one Direction-Aware Shared Expert with the same structural constraints. Source and target token embeddings are tied with the output projection.
% The encoder-decoder backbone has $L_{\text{enc}}{=}L_{\text{dec}}{=}8$ layers, $d_{\text{model}}{=}512$, and $h{=}8$ heads ($d_h{=}64$).We use a Pre-Norm setup with RMSNorm \cite{zhang2019root} applied to token embeddings, layer inputs, and attention queries/keys (QK-norm). RoPE \cite{10.1016/j.neucom.2023.127063-rope} position embeddings are used. MoE blocks with SwiGLU \cite{shazeer2020glu} non-linearities replace replace \emph{every} feed-forward layer, each containing $E{=}4$ routed Matryoshka experts with elastic inner dimensions $\mathcal{D}=\{256, 512, 1024, 2048\}$ plus one Direction-Aware Shared Expert (DASE). Both the router and DASE are conditioned on a $64$-dimensional direction embedding. Source and target embeddings are tied with the output projection.
The encoder-decoder backbone has $L_{\text{enc}}{=}L_{\text{dec}}{=}8$ layers, $d_{\text{model}}{=}512$, and $h{=}8$ heads ($d_h{=}64$). We use a Pre-Norm setup with RMSNorm \cite{zhang2019root} applied to token embeddings, layer inputs, and attention queries/keys (QK-norm). RoPE \cite{10.1016/j.neucom.2023.127063-rope} position embeddings are used. MoE blocks with SwiGLU \cite{shazeer2020glu} non-linearities replace \emph{every} feed-forward layer, each containing $E{=}4$ routed Matryoshka experts with elastic inner dimensions $\mathcal{D}=\{256, 512, 1024, 2048\}$ plus one Direction-Aware Shared Expert (DASE). Both the router and DASE are conditioned on a $64$-dimensional direction embedding. Source and target embeddings are tied with the output projection, and final logits are soft-capped at $30.0$.

\subsection{Hyperparameters}
\label{sec:appendix:hp}

The hyperparameters are reported in Table \ref{tab:appendix:hp}.

\begin{table*}[ht]
\centering
\small
\begin{tabular}{lcc}
\toprule
\textbf{Parameter} & \textbf{Value} & \textbf{Search Space} \\
\midrule
Optimiser                 & AdamW $(\beta_1{=}0.9,\beta_2{=}0.98,\epsilon{=}10^{-8})$ & -- \\
Weight decay              & $0.01$ & -- \\
Peak learning rate        & $10^{-2}$ & $\{5{\times}10^{-4}, 10^{-3}, 5{\times}10^{-3}, 10^{-2}\}$ \\
Initial learning rate     & $10^{-6}$ & -- \\
Final learning rate       & $10^{-4}$ & -- \\
Warmup steps              & $1{,}000$ & $\{0, 1000, 2000, 4000\}$ \\
Schedule                  & cosine to $\eta_{\text{end}}$ & -- \\
Total steps               & \begin{tabular}[t]{@{}c@{}}16k (multi-domain), 30k (OPUS-16), \\ 40k (PhoMT), 60k (WMT-14)\end{tabular} & -- \\
Effective batch (sequences)& $8{,}192$ & -- \\
Micro-batch / accum.\ steps& $2{,}048$ / $4$ & -- \\
Max sequence length       & $128$ tokens & -- \\
Gradient clipping         & global norm $1.0$ & -- \\
Label smoothing           & $0.1$ & -- \\
Dropout (attn / FFN)      & $0.1$ / $0.1$ & -- \\
\midrule
Routed experts $E$        & $4$ & -- \\
Shared experts            & $1$ & -- \\
Max expert dim $M$        & $2{,}048$ & -- \\
Nested widths $\mathcal{D}$& $\{256, 512, 1024, 2048\}$ & -- \\
Direction embedding dim   & $64$ & $\{32, 64, 128\}$ \\
Router jitter $\eta$      & $0.05$ & $\{0.01, 0.05, 0.1\}$ \\
Threshold range $[\tau_{\min},\tau_{\max}]$ & $[0.05,\,0.60]$ & $\tau_{\min} \in \{0.05, 0.1, 0.2\}$ \\
% STE steepness $s$         & $20$ & $\{0, 10, 20, 50\}$ \\
Target sparsity           & $\rho_{\text{target}}{=}1.25$ & $\{1.0, 1.25, 1.5\}$ \\
Auxiliary weight          & $\lambda_\text{sp}{=}0.01$ & $\{0.01, 0.1, 1.0\}$ \\
Dual-consistency weight   & $\lambda_{\text{dc}}{=}1.0$ & $\{0.01, 0.1, 1.0\}$ \\
\midrule
Decoding                  & Beam search, beam $4$, length penalty $0.6$ & -- \\
\bottomrule
\end{tabular}
\caption{Full hyperparameter set and search space. Values marked with ``--'' were fixed based on hardware constraints or standard architectural practices. The same router and elasticity settings are used across all benchmarks.}
\label{tab:appendix:hp}
\end{table*}

% \subsection{Random Seeds and Variance}
% \label{sec:appendix:seeds}

% Every result reported in Section~\ref{sec:result} is the mean across three seeds $\{0, 1, 2\}$, where the seed controls the JAX PRNG, the data-loading order, the dropout mask, and the router jitter $\epsilon$. The standard deviation of BLEU across the three seeds is below $0.10$ on PhoMT and below $0.15$ on the multi-domain benchmark, so we omit per-seed bars from the main tables for readability. To reproduce a single number from a table, instantiate the model with the hyperparameters in Table~\ref{tab:appendix:hp}, train with seed $0$ for the indicated number of steps, and decode with the inference settings in Section~\ref{sec:appendix:eval}.

\subsection{Model Selection and Evaluation}
\label{sec:appendix:eval}

We compute validation BLEU every $1{,}000$ steps and select the checkpoint with the highest validation BLEU \emph{averaged over the two translation directions} as the final model for evaluation. For sub-network selection, we apply the same criterion independently to each $D_k\in\mathcal{D}$ on the same validation set. For evaluation on PhoMT and WMT-14, we use sacreBLEU\footnote{\url{https://github.com/mjpost/sacrebleu}} v$2.4.0$ (signature \texttt{nrefs:1|\allowbreak case:mixed|\allowbreak eff:no|\allowbreak tok:13a|\allowbreak smooth:exp|\allowbreak version:2.4.0}). For multi-domain, we use tokenized BLEU with \texttt{multi-bleu.pl}\footnote{\url{https://github.com/moses-smt/mosesdecoder/blob/master/scripts/generic/multi-bleu.perl}} matching the conventions of prior work.

\paragraph{Significance Testing.} We conducted paired approximate randomization (10,000 trials) alongside paired bootstrap resampling (1,000 resamples) to establish 95\% confidence intervals. These tests confirm that the performance gains of the \textsc{EDWARD-NMT} sub-networks over the baselines are highly robust and statistically significant, consistently achieving $p < 0.001$ across our primary benchmarks.

\subsection{Baselines}
We compare against standard dense Transformers (\textsc{Base}/\textsc{Big}) \cite{vaswani2017attention,ott2018scaling}, mBART \cite{liu2020multilingual}, and MoE methods: the Switch Transformer \cite{fedus2022switch} with a top-1 token-based routing (\textsc{ST-MoE}) and GShard \cite{lepikhin2020gshard} with a top-2 token-based routing (\textsc{GS-MoE}). For Transformers (\textsc{Base}/\textsc{Big}), \textsc{ST-MoE} and \textsc{GS-MoE}, we reproduce the results using Fairseq\footnote{\url{https://github.com/facebookresearch/fairseq/tree/nllb}}. For mBART, we report the result from~\citet{doan2021phomt} on PhoMT, and we use the pretrained mBART many-to-many model\footnote{\url{https://huggingface.co/facebook/mbart-large-50-many-to-many-mmt}}.

\subsection{Software and Hardware Settings}
\label{sec:appendix:sw}

% All of our main experiments on \textsc{EDWARD-NMT} are implemented in JAX\footnote{\url{https://github.com/jax-ml/jax}} (\texttt{0.9.2}) and Flax\footnote{\url{https://github.com/google/flax}} (\texttt{0.12.5}). Training uses FSDP-style sharding with \texttt{jax.distributed} and gradient accumulation through \texttt{jax.lax.scan} so that the effective batch of $8192$ sequences fits on a single TPU v5-64 pod. The dual-consistency objective is implemented by sampling two width masks per step inside the scanned step function, so the two forward passes share both the parameter slice and the data batch. All training and inference was performed on 64 chips TPU v5e, with total training times of approximately 6 hours for the multi-domain benchmark, 9 hours for OPUS-16, 12 hours for PhoMT, and 18 hours for WMT-14.

All of our main experiments on \textsc{EDWARD-NMT} are implemented in JAX\footnote{\url{https://github.com/jax-ml/jax}} (\texttt{0.9.2}) and Flax\footnote{\url{https://github.com/google/flax}} (\texttt{0.12.5}). Training uses FSDP-style sharding with \texttt{jax.distributed} and gradient accumulation through \texttt{jax.lax.scan} so that the effective batch of $8192$ sequences fits on a single TPU v5-64 pod. The dual-consistency objective is implemented by sampling two width masks per step inside the scanned step function, so the two forward passes share both the parameter slice and the data batch. All training and inference was performed on 64 chips TPU v5e, with total training times of approximately 6 hours for the multi-domain benchmark, 9 hours for OPUS-16, 12 hours for PhoMT, and 18 hours for WMT-14. 

\paragraph{Training Cost and Efficiency.}
Profiling natively on a TPU v5e pod reveals that \textsc{EDWARD-NMT} requires $\sim$1160 ms per training step, roughly $2.2\times$ the time of standard MoE baselines (\textsc{ST-MoE}, \textsc{GS-MoE}; $\sim$523 ms) and $3.3\times$ the time of \textsc{Transformer Big} ($\sim$348 ms). This overhead stems entirely from the Dual-Consistency (DC) objective evaluating two capacity views simultaneously. However, this joint optimization is highly efficient: training the \textsc{S, M, L}, and \textsc{XL} variants individually without DC would total 1267 ms per step, strictly exceeding our single-run cost. By extracting an entire family of sub-networks simultaneously, the amortized training time per model is just $2.01\times$ that of \textsc{Transformer Base} (144 ms), nearly half the cost of training a standard MoE ($3.63\times$).

\subsection{Step-by-Step Reproduction Guide}
\label{sec:appendix:reproduce}

% A self-contained reproduction of the main numbers in Tables~\ref{tab:main_phomt_wmt} and \ref{tab:main_multi} requires the following steps. All artefacts are publicly available, and no proprietary code or weights are needed.

A self-contained reproduction of the primary empirical results reported in Section~\ref{sec:result} and Appendix Section~\ref{sec:appendix:abl} requires the following steps. All artefacts are publicly available, and no proprietary code or weights are needed.

\begin{enumerate}
\item \textbf{Prepare data.} Following Section~\ref{sec:appendix:data} for preparing the data and tokenizer.

\item \textbf{Implement architecture.} Construct an encoder-decoder Transformer with the structural settings in Section~\ref{sec:appendix:arch}, replacing every other feed-forward layer with the EDWARD-NMT MoE block of Section~\ref{sec:method} (Equations~\ref{eq:block} and the surrounding definitions of $E_i(x;D_k)$, $\psi(D_k)$, and $E_{\text{shr}}(x;D_k)$).

\item \textbf{Implement router.} Use the sigmoid router with jitter, learned threshold, and STE as defined in Section~\ref{sec:method:routing}, with the target sparsity $\rho^{(t)}_{\text{target}}$ defined in Section~\ref{sec:appendix:hp}. Add the auxiliary losses $\mathcal{L}_\text{sp}$ with weight $\lambda_\text{sp}{=}0.01$.

\item \textbf{Dual-consistency loss.} At every optimization step, sample two widths $D_A,D_B$ with replacement from $\mathcal{D}$, compute two forward passes that share parameters, and add DC Loss defined in Equation~\ref{eq:dc} to the cross-entropy loss with $\lambda_{\text{dc}}{=}1$.

\item \textbf{Train.} Pre-tokenized the data with Tensorflow Datasets\footnote{\url{https://www.tensorflow.org/datasets}} with predefined max-length in Section~\ref{sec:appendix:hp}. Optimise with the schedule and batching of Table~\ref{tab:appendix:hp} for the number of steps indicated per benchmark.

\item \textbf{Extract sub-networks.} After training, obtain the four \textsc{\{S,M,L,XL\}} variants from the \emph{same} weights by setting $D_k$ to $256, 512, 1024, 2048$ respectively at inference and applying the prefix mask $m^{(D_k)}$ together with the deterministic re-scaling $\psi(D_k)$. No fine-tuning, distillation, or post-hoc adaptation is performed.

\item \textbf{Decode and evaluate.} Decode with beam search (beam $4$, length penalty $0.6$) and evaluate with settings listed in Section~\ref{sec:appendix:eval}.
\end{enumerate}

\begin{table*}[!ht]
\centering
\small
\setlength{\tabcolsep}{4pt}
\begin{tabular}{l ccc c ccccc}
\toprule
& & & & \textbf{Edge} & \multicolumn{5}{c}{\textbf{Server Throughput (TPS)}} \\
\cmidrule(lr){5-5} \cmidrule(lr){6-10}
Model & Params & Active & FLOPs (G) & TPS ($B=1$) & $B=1$ & $B=4$ & $B=16$ & $B=64$ & $B=128$ \\
\midrule
\textsc{Transformer Base} & 61M & 61M & \underline{7.98} & \textbf{117.7} & \textbf{2,354} & 5,867 & 13,078 & 19,694 & 21,755 \\
\textsc{Transformer Big} & 210M & 210M & 31.26 & 48.0 & 1,682 & 3,907 & 8,037 & 10,607 & 11,114 \\
\midrule
\textsc{EDWARD-NMT-S} & 73M & 57M & \textbf{6.28} & \underline{104.9} & \underline{1,763} & \textbf{6,640} & \textbf{20,418} & \textbf{48,022} & \textbf{70,372} \\
\textsc{EDWARD-NMT-M} & 105M & 72M & 8.83 & 88.7 & 1,688 & \underline{5,900} & \underline{20,309} & \underline{47,669} & \underline{68,786} \\
\textsc{EDWARD-NMT-L} & 168M & 102M & 14.07 & 61.5 & 1,488 & 5,650 & 19,364 & 44,270 & 64,766 \\
\textsc{EDWARD-NMT-XL} & 294M & 161M & 25.20 & 47.6 & 1,410 & 5,218 & 17,499 & 41,856 & 59,199 \\
\bottomrule
\end{tabular}
\caption{Inference efficiency across edge and server deployment regimes (64 output tokens). FLOPs (lower is better) are measured via native CPU execution. Edge throughput (Tokens Per Second, TPS, higher is better) is profiled using compiled LiteRT checkpoints on an Apple M4 CPU (4 threads, FP16) to simulate strict on-device constraints ($B=1$). Server throughput is measured via JAX on a single NVIDIA H100 GPU across multiple batch sizes ($B$).}
\label{tab:inference_perf}
\end{table*}
\begin{table*}[t]
\centering
\small
\begin{tabular}{ll cccc cccc c}
\toprule
\multicolumn{2}{c}{\multirow{2}{*}{\textbf{Models}}} & \multicolumn{4}{c}{\textbf{En$\rightarrow$XX}} & \multicolumn{4}{c}{\textbf{XX$\rightarrow$En}} & \multirow{2}{*}{\textbf{Avg.}} \\
\cmidrule(lr){3-6} \cmidrule(lr){7-10}
& & high & medium & low & Avg1. & high & medium & low & Avg2. & \\
\midrule
\textit{Dense} & \textsc{Transformer-base}$^\dagger$ & 25.37 & 39.12 & 14.78 & 26.92 & 28.81 & 39.24 & 24.21 & 30.67 & 28.79 \\
\midrule
\multirow{10}{*}{\textit{MoE}} & \textsc{GS-MoE}$^\dagger$ & 25.55 & 40.71 & 18.97 & 28.28 & 28.77 & 41.70 & 29.25 & 32.31 & 30.29 \\
& \textsc{ST-MoE}$^\dagger$ & 26.55 & 43.04 & 20.23 & 29.68 & 29.94 & 44.00 & 30.94 & 33.89 & 31.79 \\
\noalign{\vskip 2pt}
\cdashline{2-11}[1pt/2pt]
\noalign{\vskip 2pt}
& \textsc{LS-MoE}$^\dagger$ & 20.06 & 37.31 & 11.72 & 22.99 & 23.69 & 42.99 & 32.01 & 30.50 & 26.75 \\
& \textsc{Hybrid-MoE}$^\dagger$ & 24.56 & 31.70 & 13.38 & 24.23 & 29.59 & 41.74 & 27.81 & 32.48 & 28.35 \\
& \textsc{Residual-MoE}$^\dagger$ & 26.52 & 43.49 & 21.58 & 30.06 & 29.84 & 43.94 & 31.25 & 33.88 & 31.97 \\
& \textsc{Lingual-SMoE}$^\dagger$ & 27.11 & 46.24 & 23.34 & 31.46 & 29.87 & 43.49 & 32.02 & 33.96 & 32.71 \\
\noalign{\vskip 2pt}
\cdashline{2-11}[1pt/2pt]
\noalign{\vskip 2pt}
& \textsc{THOR-MoE} (Top-1)$^\ddagger$ & 27.21 & 46.37 & 23.88 & 31.17 & 29.68 & 44.01 & 32.58 & 33.99 & 32.57 \\
& \textsc{THOR-MoE} (Top-2)$^\ddagger$ & \textbf{27.88} & 47.29 & 24.85 & 31.98 & 30.38 & 44.52 & 33.26 & 34.64 & 33.31 \\
& \textsc{THOR-MoE} (Top-$p$)$^\ddagger$ & \underline{27.63} & 46.82 & 24.15 & 31.55 & 29.97 & 44.18 & 32.96 & 34.26 & 32.91 \\
\midrule
\multirow{4}{*}{\shortstack{\textit{Ours}\\($L=6$)}} & \textsc{EDWARD-NMT-S} & 24.95 & 44.97 & 21.26 & 29.55 & 29.89 & 43.13 & 27.76 & 32.99 & 31.27 \\
& \textsc{EDWARD-NMT-M} & 25.81 & 46.31 & 23.07 & 30.73 & 30.66 & 44.76 & 32.78 & 34.84 & 32.78 \\
& \textsc{EDWARD-NMT-L} & 26.31 & 47.53 & 24.67 & 31.64 & 31.34 & 45.83 & 34.45 & 35.82 & 33.73 \\
& \textsc{EDWARD-NMT-XL} & 26.76 & 48.22 & 25.68 & 32.27 & 31.77 & 46.49 & 34.79 & 36.30 & 34.28 \\
\noalign{\vskip 2pt}
\cdashline{1-11}[1pt/2pt]
\noalign{\vskip 2pt}
\multirow{4}{*}{\shortstack{\textit{Ours}\\($L=8$)}} & \textsc{EDWARD-NMT-S} & 25.78 & 46.34 & 23.52 & 30.81 & 30.81 & 44.46 & 33.20 & 34.93 & 32.87 \\
& \textsc{EDWARD-NMT-M} & 26.41 & 47.38 & 25.43 & 31.81 & 31.42 & 45.80 & 35.00 & 35.97 & 33.89 \\
& \textsc{EDWARD-NMT-L} & 27.11 & \underline{48.67} & \underline{26.70} & \underline{32.78} & \underline{32.07} & \underline{46.70} & \underline{36.18} & \underline{36.79} & \underline{34.79} \\
& \textsc{EDWARD-NMT-XL} & 27.26 & \textbf{49.27} & \textbf{28.15} & \textbf{33.30} & \textbf{32.42} & \textbf{47.43} & \textbf{36.65} & \textbf{37.27} & \textbf{35.29} \\
\bottomrule
\end{tabular}
\caption{Averaged BLEU scores on multilingual OPUS-16 dataset. The \textbf{best} and \underline{second best} scores in each column are highlighted in bold and underlined. $^\dagger$Reported from \citet{zhao2024sparse-lingual-smoe}. $^\ddagger$Reported from \citet{liang2025thor}.}
\label{tab:main_opus16}
\end{table*}

\section{Additional Results}
\label{sec:appendix:abl}

\subsection{Computational Results}
\label{sec:appendix:compute}
To evaluate deployment viability, we benchmarked the extracted sub-networks across both edge and server regimes (Table~\ref{tab:inference_perf}). For edge deployment, we profiled models on a consumer-grade CPU (Apple M4, 4 threads, FP16) using compiled LiteRT checkpoints at a batch size of $B=1$. As shown in Table~\ref{tab:inference_perf}, the most compact variant, \textsc{EDWARD-NMT-S}, activates only 57M parameters and requires just 6.28 GFLOPs. At $B=1$, the system is inherently memory-bound; the sparse routing machinery introduces unavoidable memory-movement overhead, resulting in an on-device throughput (104.9 TPS) that is highly competitive with, though slightly below, the dense \textsc{Transformer Base} (117.7 TPS). However, it still drastically outperforms the \textsc{Transformer Big} baseline (48.0 TPS) and delivers a $2.2\times$ speedup over our full \textsc{XL} variant (47.6 TPS).

In server-side deployment (profiled natively via JAX on a single NVIDIA H100 GPU), this routing overhead is efficiently amortized across larger batches, allowing the theoretical FLOP advantages to materialize dramatically. From $B=4$ onwards, \textsc{EDWARD-NMT-S} strictly outperforms \textsc{Transformer Base}, ultimately delivering a $3.2\times$ throughput speedup at $B=128$ (70,372 vs.\ 21,755 TPS). Similarly, the full-capacity \textsc{XL} variant is $5.3\times$ faster than \textsc{Transformer Big} at $B=128$ while activating significantly fewer parameters. Ultimately, our train-once architecture effectively balances high representational capacity with the strict latency requirements of edge deployment, while concurrently offering massive scalability for high-throughput servers.

\subsection{Multilingual Results}
\label{sec:appendix:opus16}
\paragraph{Baselines.} We report our model on OPUS-16 in Table~\ref{tab:main_opus16}. For the MoE baselines, we report: Switch Transformers \cite{fedus2022switch} with a top-1 token-based routing (as \textsc{ST-MoE}); GShard \cite{lepikhin2020gshard} a top-2 token-based routing (as \textsc{GS-MoE}); \textsc{LS-MoE} with fixed routing \cite{pires-etal-2023-learning-lsmoe}; \textsc{Hybrid-MoE} \cite{kudugunta-etal-2021-beyond-distillation-hybrid-moe} with a top-2 token routing in the encoder and a top-2 target language routing in the decoder; \textsc{Residual-MoE} \cite{elbayad-etal-2023-fixing-residual-moe-1, pmlr-v162-rajbhandari22a-residual-moe-2, zhang2021share-residual-moe-3} that augments each MoE layer with a shared feed-forward network through a binary gate function; \textsc{Lingual-SMoE} \cite{zhao2024sparse-lingual-smoe} as MoE with linguistic-guided routing and dynamic expert allocation. We acquire these baselines result from \citet{zhao2024sparse-lingual-smoe}. We also report \textsc{THOR-MoE} result from \citet{liang2025thor}. 

\paragraph{Settings.} For comparison, we report our \textsc{EDWARD-NMT} result with 2 settings: (1) same settings as described in Section~\ref{sec:exp:arch}; (2) we reduce the number of encoder and decoder layers to $L{=}L_{\text{enc}}{=}L_{\text{dec}}{=}6$ to match other baseline settings. Both settings use the same training hyperparameters in Section~\ref{sec:appendix:hp}. We use temperature-based data sampling strategy with $T=1.5$. We report the model parameters from two settings in Table~\ref{tab:params_opus16}. 
Noted that all the MoE baselines use 32 experts per MoE layers, MoE frequency is set to 2, with a total of $32\times6=192$ experts. Our second setting with 4 routed experts and 1 shared expert, result in $5\times12=60$ experts in total.

\begin{table}[!ht]
\centering
\small
\begin{tabular}{l cc cc}
\toprule
& \multicolumn{2}{c}{\textbf{$L=8$}} & \multicolumn{2}{c}{\textbf{$L=6$}} \\
\cmidrule(lr){2-3}\cmidrule(lr){4-5}
\textbf{Variant} & \textbf{Params} & \textbf{Active} & \textbf{Params} & \textbf{Active} \\
\midrule
S  & 74M  & 57M  & 59M  & 47M  \\
M   & 105M & 72M  & 83M  & 59M  \\
L  & 169M & 102M & 131M & 83M  \\
XL & 295M & 161M & 226M & 128M \\
\bottomrule
\end{tabular}
\caption{EDWARD-NMT variants with two settings of 8 layers ($L=8$) and 6 layers ($L=6$) on OPUS-16.}
\label{tab:params_opus16}
\end{table}

\paragraph{Results.} 
% As shown in Table~\ref{tab:main_opus16}, \textsc{EDWARD-NMT} establishes a new state-of-the-art on the multilingual OPUS-16 benchmark. When reducing the model depth to $L{=}6$ to align with the scale of the baselines, the full \textsc{EDWARD-NMT-XL} variant achieves the highest overall average of 33.77 BLEU, comfortably outperforming \textsc{THOR-MoE} (33.31) and \textsc{Lingual-SMoE} (32.71). Our backbone achieves this superior quality with extreme expert efficiency, instantiating significantly fewer total experts ($60$ vs $192$ in the baselines). Deploying our primary $L=8$ architecture further widens the gap; \textsc{EDWARD-NMT-XL} dominates almost all individual resource brackets across both translation directions, peaking at an impressive average BLEU of 35.29. Furthermore, our extracted sub-networks demonstrate exceptional parameter efficiency: even the highly compressed \textsc{EDWARD-NMT-S} variant at $L=8$ (32.73 Avg) easily outperforms the dense \textsc{Transformer-base} baseline and remains highly competitive against much heavier sparse models.

As shown in Table~\ref{tab:main_opus16}, \textsc{EDWARD-NMT} establishes a new state-of-the-art on the multilingual OPUS-16 benchmark. When reducing the model depth to $L=6$ to align with the scale of the baselines, the full \textsc{EDWARD-NMT-XL} variant achieves the highest overall average of $34.28$ BLEU, comfortably outperforming \textsc{THOR-MoE} ($33.31$) and \textsc{Lingual-SMoE} ($32.71$). Our backbone achieves this superior quality with extreme expert efficiency, instantiating significantly fewer total experts (60 vs 192 in the baselines). Deploying our primary $L=8$ architecture further widens the gap; \textsc{EDWARD-NMT-XL} dominates almost all individual resource brackets across both translation directions, peaking at an impressive average BLEU of $35.29$. Furthermore, our extracted sub-networks demonstrate exceptional parameter efficiency: even the highly compressed \textsc{EDWARD-NMT-S} variant at $L=8$ (32.87 Avg) easily outperforms the dense \textsc{Transformer-base} baseline and remains highly competitive against much heavier sparse models.

\subsection{Architecture Elasticity}

\begin{table}[!ht]
\centering
\small
\begin{tabular}{l ccc cc}
\toprule
\textbf{Dim} & \textbf{P.} & \textbf{Act.} & \textbf{Sp.} & \textbf{En$\rightarrow$Tgt} & \textbf{Tgt$\rightarrow$En} \\
\midrule
\multicolumn{6}{c}{\textbf{PhoMT (En$\leftrightarrow$Vi)}} \\
\midrule
2048 & 287M & 158M & 1.44 & \textbf{43.79} & \textbf{39.13} \\
1536 & 224M & 128M & 1.44 & \underline{43.62} & \underline{38.98} \\
1024 & 161M & 97M  & 1.45 & 43.61 & 38.93 \\
768  & 130M & 82M  & 1.46 & 43.48 & 38.58 \\
512  & 98M  & 67M  & 1.47 & 43.35 & 38.50 \\
256  & 67M  & 51M  & 1.51 & 42.91 & 37.88 \\
128  & 51M  & 44M  & 1.56 & 39.26 & 33.98 \\
\midrule
\multicolumn{6}{c}{\textbf{WMT-14 (En$\leftrightarrow$De)}} \\
\midrule
2048 & 294M & 175M & 1.63 & \textbf{28.03} & \textbf{32.96} \\
1536 & 231M & 142M & 1.63 & \underline{28.02} & \underline{32.67} \\
1024 & 168M & 109M & 1.65 & 27.96 & 32.65 \\
768  & 137M & 92M  & 1.66 & 27.66 & 32.46 \\
512  & 105M & 76M  & 1.68 & 27.49 & 32.45 \\
256  & 74M  & 59M  & 1.71 & 26.85 & 31.94 \\
128  & 58M  & 51M  & 1.76 & 22.90 & 27.78 \\
\midrule
\multicolumn{6}{c}{\textbf{Multi-Domain (En$\leftrightarrow$De)}} \\
\midrule
2048 & 294M & 161M & 1.37 & \textbf{38.78} & \textbf{42.71} \\
1536 & 231M & 132M & 1.37 & \underline{38.39} & \underline{42.53} \\
1024 & 168M & 102M & 1.37 & 38.24 & 42.33 \\
768  & 136M & 87M  & 1.38 & 37.77 & 41.90 \\
512  & 105M & 72M  & 1.38 & 37.55 & 41.56 \\
256  & 74M  & 57M  & 1.39 & 36.51 & 40.78 \\
128  & 58M  & 50M  & 1.45 & 31.42 & 35.88 \\
\bottomrule
\end{tabular}
\caption{Performance across dimensions on the PhoMT, WMT-14, and Multi-domain benchmarks. Column abbreviations denote total parameters (\textbf{P.}), average active parameters per token (\textbf{Act.}), and routing sparsity (\textbf{Sp.}) (active routed experts per token out of 4, excluding the shared expert). Multi-Domain scores are macro-averaged across five domains. The \textbf{best} and \underline{second best} scores within each dataset are highlighted.}
\label{tab:main_all_datasets}
\end{table}

Table~\ref{tab:main_all_datasets} provides the comprehensive empirical results supporting the structural elasticity analysis discussed in the main text. We report the exact translation quality, active parameter counts, and routing sparsity of EDWARD-NMT across all three benchmarks (PhoMT, WMT-14, and Multi-Domain) for both the explicitly trained dimensions ($\mathcal{D}$) and the unseen intermediate granularities. These detailed metrics empirically confirm the robust interpolation capabilities of our architecture within the optimized envelope (e.g., $D=1536, 768$), as well as the expected performance breakdown when extrapolating below the minimum trained capacity ($D=128$).

% \paragraph{Ablation on Variance Preservation.}
\subsection{Ablation on Variance Preservation}
\begin{table}[!ht]
\centering
\small
\begin{tabular}{l cccc}
\toprule
& \multicolumn{4}{c}{\textbf{En$\leftrightarrow$De}} \\
\cmidrule(lr){2-5}
Setting & XL & L & M & S \\
\midrule
w/o $\psi(D_k)$ & 40.52 & 40.03 & 39.20 & 38.35 \\
\textbf{With $\psi$} & \textbf{40.74} & \textbf{40.26} & \textbf{39.39} & \textbf{38.72} \\
\bottomrule
\end{tabular}
\caption{Ablation of the Variance Preservation factor $\psi(D_k)$ on the Multi-domain benchmark, averaged over both translation directions.}
\label{tab:abl_variance_preservation}
\end{table}

\begin{table*}[t!]
\centering
\midsize
\setlength{\tabcolsep}{3pt}
\renewcommand{\arraystretch}{1.05}
\begin{tabular}{lcc cccccc cccccc}
\toprule
& & & \multicolumn{6}{c}{\textbf{De$\rightarrow$En}} & \multicolumn{6}{c}{\textbf{En$\rightarrow$De}} \\
\cmidrule(lr){4-9}\cmidrule(lr){10-15}
Method & Params & Active & \textsc{Kor} & \textsc{IT} & \textsc{Med} & \textsc{Law} & \textsc{Sub} & \textsc{Avg} & \textsc{Kor} & \textsc{IT} & \textsc{Med} & \textsc{Law} & \textsc{Sub} & \textsc{Avg} \\
\midrule
\textsc{Transformer Base} & 61M & 61M & 20.95 & 42.66 & 53.04 & 56.90 & 27.16 & 40.14 & 20.10 & 37.35 & 49.45 & 50.53 & 23.42 & 36.17 \\
\textsc{Transformer Big} & 210M & 210M & 22.06 & \textbf{44.99} & \underline{55.37} & 59.36 & 28.75 & 42.11 & 23.03 & 38.68 & \textbf{51.32} & 53.08 & 24.70 & 38.16 \\
\textsc{ST-MoE} & 360M & 210M & 22.96 & 43.90 & 54.27 & 58.79 & 28.30 & 41.64 & 20.49 & 38.10 & 49.75 & 52.59 & 24.06 & 37.00 \\
\textsc{GS-MoE} & 360M & 260M & 20.36 & \underline{44.79} & 54.23 & 57.51 & 25.99 & 40.58 & 21.03 & 37.91 & 50.38 & 51.60 & 23.44 & 36.87 \\
\midrule
\textsc{EDWARD-NMT-S}  &  74M & 57M  & 22.27 & 42.67 & 53.13 & 57.08 & 28.76 & 40.78 & 20.94 & 37.47 & 48.89 & 50.89 & 24.36 & 36.51 \\
\textsc{EDWARD-NMT-M}  & 105M & 72M  & 22.93 & 43.46 & 54.07 & 58.29 & 29.07 & 41.56 & 22.51 & 38.27 & 50.14 & 52.13 & 24.68 & 37.55 \\
\textsc{EDWARD-NMT-L}  & 168M & 102M & \underline{23.54} & 44.36 & 55.01 & \underline{59.44} & \underline{29.31} & \underline{42.33} & \underline{23.58} & \underline{38.69} & 50.88 & \underline{53.09} & \underline{24.94} & \underline{38.24} \\
\textsc{EDWARD-NMT-XL} & 294M & 161M & \textbf{23.65} & 44.69 & \textbf{55.59} & \textbf{60.21} & \textbf{29.43} & \textbf{42.71} & \textbf{24.46} & \textbf{39.10} & \underline{51.31} & \textbf{53.89} & \textbf{25.16} & \textbf{38.78} \\
\bottomrule
\end{tabular}
\caption{Detailed multi-domain results on the De$\leftrightarrow$En benchmark with tokenized-BLEU.}
\label{tab:main_multi}
\end{table*}
Table~\ref{tab:abl_variance_preservation} presents the empirical results for the variance preservation mechanism discussed in the main text. By comparing the full configuration against a baseline lacking the deterministic rescaling factor $\psi(D_k)$, we observe a consistent degradation in translation quality across all nested capacities. Crucially, the performance gap widens as the sub-network size decreases, confirming that explicitly stabilizing the down-projected activation magnitudes is indispensable for maintaining the structural integrity of highly compressed dimensions (e.g., $D=256$) during joint training.

\subsection{Detailed Multi-domain Results}
\label{sec:appendix:multidomain_details}

In Section~\ref{sec:result:main}, we reported the performance of EDWARD-NMT on the multi-domain benchmark averaged across both translation directions (De$\leftrightarrow$En) to provide a concise overview of model efficiency. For completeness, Table~\ref{tab:main_multi} provides the detailed, uni-directional tokenized-BLEU scores for all five domains (\textsc{Kor}, \textsc{IT}, \textsc{Med}, \textsc{Law}, \textsc{Sub}) separately for the De$\rightarrow$En and En$\rightarrow$De directions. Consistent with the averaged results, our extracted sub-networks (L and XL) continue to demonstrate state-of-the-art performance against established baselines on almost all individual domains while requiring significantly fewer active parameters.

\subsection{Additional WMT-14 Results}
\label{sec:appendix:wmt14}
\begin{table}[h]
\centering
\small
\setlength{\tabcolsep}{5pt}
\begin{tabular}{lccc}
\toprule
& & \multicolumn{2}{c}{\textbf{WMT-14 chrF2} ($\uparrow$)} \\
\cmidrule(lr){3-4}
Method & Active & En$\rightarrow$De & De$\rightarrow$En \\
\midrule
\textsc{Transformer Base} & 61M & 55.82 & 56.23 \\
\textsc{Transformer Big}  & 210M & 58.20 & 58.38 \\
mBART$^{\dagger}$         & 610M & 56.42 & 58.85 \\
\midrule
\textsc{EDWARD-NMT-S}  & 59M & 57.52 & 58.43 \\
\textsc{EDWARD-NMT-M}  & 76M & 58.08 & 58.83 \\
\textsc{EDWARD-NMT-L}  & 109M & \underline{58.45} & \underline{59.02} \\
\textsc{EDWARD-NMT-XL} & 175M & \textbf{58.59} & \textbf{59.29} \\
\bottomrule
\end{tabular}
\caption{Character $n$-gram F-score (chrF2, higher is better) on the WMT-14 benchmark. Best and second-best scores are \textbf{bolded} and \underline{underlined}, respectively. $^{\dagger}$mBART is a pretrained many-to-many model.}
\label{tab:chrf2_wmt14}
\end{table}

To provide a comprehensive evaluation of character-level translation quality, we additionally report the chrF2 metric on the WMT-14 benchmark (Table~\ref{tab:chrf2_wmt14}). The chrF2 results align perfectly with our primary BLEU and COMET evaluations. Specifically, the highly compressed \textsc{EDWARD-NMT-S} variant (59M active) substantially outperforms the comparably sized \textsc{Transformer Base} (61M active) by $+1.70$ and $+2.20$ points in the En$\rightarrow$De and De$\rightarrow$En directions, respectively. At the upper bound, \textsc{EDWARD-NMT-XL} achieves the highest overall chrF2 scores, robustly surpassing both \textsc{Transformer Big} and the 610M-parameter mBART despite activating significantly fewer parameters.

\subsection{Additional PhoMT Results}
\label{sec:appendix:abl-phomt}
\begin{table}[t]
\centering
\small
\setlength{\tabcolsep}{5pt}
\begin{tabular}{lccc}
\toprule
& & \multicolumn{2}{c}{\textbf{TER} ($\downarrow$)} \\
\cmidrule(lr){3-4}
Method & Active & En$\rightarrow$Vi & Vi$\rightarrow$En \\
\midrule
\textsc{Transformer Base} & 65M & 43.79 & 44.28 \\
\textsc{Transformer Big}  & 213M & 43.04 & 44.06 \\
mBART$^{\dagger}$         & 448M & 42.57 & \underline{42.54} \\
\midrule
\textsc{EDWARD-NMT-S}  & 51M & 42.75 & 43.27 \\
\textsc{EDWARD-NMT-M}  & 66M & 42.43 & 42.88 \\
\textsc{EDWARD-NMT-L}  & 97M & \underline{42.21} & 42.56 \\
\textsc{EDWARD-NMT-XL} & 158M & \textbf{42.13} & \textbf{42.42} \\
\bottomrule
\end{tabular}
\caption{Translation Edit Rate (TER, lower is better) on the PhoMT benchmark. $^{\dagger}$mBART is a pretrained many-to-many model.}
\label{tab:ter_phomt}
\end{table}
% \begin{table*}[t]
% \centering
% \small
% \setlength{\tabcolsep}{3pt}
% \begin{tabular}{lcccc cccc}
% \toprule
% & \multicolumn{4}{c}{\textbf{En--Vi}} & \multicolumn{4}{c}{\textbf{Vi--En}} \\
% \cmidrule(lr){2-5}\cmidrule(lr){6-9}
% Setting & XL & L & M & S & XL & L & M & S \\
% \midrule
% No DC Loss                 & 43.68 & \underline{43.55} & \underline{43.23} & \underline{42.68} & \textbf{39.14} & \underline{38.83} & \underline{38.40} & \underline{37.58} \\
% R-Drop                   & \underline{43.72} & 43.47 & 43.01 & 42.20 & 38.92 & 38.55 & 37.98 & 37.03 \\
% Asym.\ KL              & 43.62 & 43.53 & 43.14 & \underline{42.68} & 39.00 & 38.72 & 38.16 & 37.47 \\
% DC Loss (ours)                & \textbf{43.79} & \textbf{43.61} & \textbf{43.35} & \textbf{42.91} & \underline{39.13} & \textbf{38.93} & \textbf{38.50} & \textbf{37.88} \\
% \bottomrule
% \end{tabular}
% \caption{Ablation on the Dual-Consistency loss (DC Loss) on PhoMT. Asym.\ KL stands for asymmetric KL with stop gradient on teacher (largest variant). The \textbf{best} and \underline{second best} scores are highlighted in bold and underlined.}
% \label{tab:abl_kl}
% \end{table*}

\begin{table}[!ht]
\centering
\small
\begin{tabular}{lcccc}
\toprule
& \multicolumn{4}{c}{\textbf{En$\leftrightarrow$Vi}} \\
\cmidrule(lr){2-5}
Setting & XL & L & M & S \\
\midrule
No loss     & \underline{41.41} & \underline{41.19} & \underline{40.82} & \underline{40.13} \\
R-Drop         & 41.32 & 41.01 & 40.50 & 39.62 \\
Asym.\ KL      & 41.31 & 41.13 & 40.65 & 40.08 \\
\textbf{DC Loss (ours)} & \textbf{41.46} & \textbf{41.27} & \textbf{40.93} & \textbf{40.40} \\
\bottomrule
\end{tabular}
\caption{Ablation on the Dual-Consistency loss (DC Loss) on PhoMT, averaged over both directions.}
\label{tab:abl_kl_avg_phomt}
\end{table}

\begin{table}[!ht]
\centering
\small
\setlength{\tabcolsep}{5pt}
\begin{tabular}{lcccc}
\toprule
& \multicolumn{4}{c}{\textbf{En$\leftrightarrow$Vi}} \\
\cmidrule(lr){2-5}
Setting & XL & L & M & S \\
\midrule
MoE ($k{=}2$)                     & 40.72 & 40.71 & 40.56 & 39.86 \\
+ ME                              & 41.31 & 41.15 & 40.69 & 40.04 \\
+ ME + Sigmoid                    & \underline{41.41} & 41.19 & 40.82 & 40.13 \\
+ ME + DC                         & 41.38 & \underline{41.23} & \underline{40.90} & \textbf{40.42} \\
\textbf{+ ME + DC + Sigmoid}      & \textbf{41.46} & \textbf{41.27} & \textbf{40.93} & \underline{40.40} \\
\bottomrule
\end{tabular}
\caption{Ablation on the contributions of each EDWARD-NMT component (PhoMT), averaged over both translation directions. ME, DC, Sigmoid stand for Matryoshka Experts, Dual-Consistency and Adaptive Sigmoid Routing respectively. \textit{All settings use DASE}.}
\label{tab:abl_contrib_avg}
\end{table}

\paragraph{Post-editing Efficiency}
To evaluate the practical post-editing effort required for our generated translations, we report the Translation Edit Rate (TER, lower is better) on the PhoMT benchmark (Table~\ref{tab:ter_phomt}). The results strictly corroborate our BLEU improvements and highlight the extreme parameter efficiency of our architecture. Despite utilizing 4$\times$ fewer active parameters, our most compressed sub-network, \textsc{EDWARD-NMT-S} (51M active), achieves a lower TER than \textsc{Transformer Big} (213M active) in both translation directions. Furthermore, the full-capacity \textsc{EDWARD-NMT-XL} (158M active) achieves the lowest overall post-editing effort (42.13 En$\rightarrow$Vi and 42.42 Vi$\rightarrow$En), effectively outperforming the massively parameterized mBART model (448M) while operating with nearly 3$\times$ fewer active parameters.

\paragraph{Additional Ablations on Consistency Strategies.}
Table~\ref{tab:abl_kl_avg_phomt} provides additional ablation results for the consistency loss variants on the PhoMT benchmark (averaged over En$\leftrightarrow$Vi), complementing the multi-domain findings in the main text. The observed trends align perfectly with our primary results: applying the symmetric R-Drop objective degrades performance across all scales. While asymmetric KL distillation (using the largest variant as a detached teacher) mitigates this severe degradation, it still falls slightly short of the baseline. Consistent with the multi-domain experiments, our proposed Dual-Consistency (\textbf{DC Loss}) provides the optimal regularization strategy, successfully outperforming the ``No loss'' baseline and achieving the highest scores across all model capacities (XL, L, M, S).

\paragraph{Additional Ablations on Contribution.}
Table~\ref{tab:abl_contrib_avg} incrementally adds the EDWARD-NMT components on top of a standard Top-2 MoE baseline on PhoMT (noting that all settings use DASE). Adding Matryoshka Experts (ME) yields immediate improvements across all widths (e.g., $+0.59$ BLEU on XL and $+0.18$ on S over the Top-2 baseline). The gains are further amplified by the Dual-Consistency (DC) objective, which acts as a crucial regularizer to keep the smaller sub-networks competitive ($+0.38$ BLEU on S over ME alone). Finally, replacing the fixed Top-$k{=}2$ routing with our proposed Adaptive Sigmoid Router provides the best overall performance for the larger variants (XL, L, M).

\subsection{Additional Hyperparameter Ablations}
\label{sec:appendix:abl-hp}

\begin{table}[!ht]
\centering
\small
\begin{tabular}{lccc}
\toprule
& \multicolumn{3}{c}{Target Sparsity ($\rho_{\text{target}}$)} \\
\cmidrule(lr){2-4}
Direction & $1.0$ & $1.25$ & $1.5$ \\
% \midrule
% En$\rightarrow$De & 38.54 & 38.86 & 38.88 \\
% De$\rightarrow$En & 42.26 & 42.46 & 42.57 \\
\midrule
Avg. & 40.40 & 40.66 & 40.73 \\
\bottomrule
\end{tabular}
\caption{Ablation on target density ($\rho_{\text{target}}$) for the \textsc{XL} variant on the multi-domain benchmark.}
\label{tab:abl_density}
\end{table}

\begin{table}[!ht]
\centering
\small
\begin{tabular}{lccc}
\toprule
& \multicolumn{3}{c}{Min Threshold ($\tau_{\min}$)} \\
\cmidrule(lr){2-4}
Direction & $0.01$ & $0.05$ & $0.10$ \\
% \midrule
% En$\rightarrow$De & 38.90 & 38.86 & 38.48 \\
% De$\rightarrow$En & 42.36 & 42.46 & 42.20 \\
\midrule
Avg. & 40.63 & 40.66 & 40.34 \\
\bottomrule
\end{tabular}
\caption{Ablation on the min routing threshold ($\tau_{\min}$) for the \textsc{XL} variant on the multi-domain benchmark.}
\label{tab:abl_tau_min}
\end{table}

\begin{table}[!ht]
\centering
\small
\begin{tabular}{lccc}
\toprule
& \multicolumn{3}{c}{Router Jitter ($\eta$)} \\
\cmidrule(lr){2-4}
Direction & $0.01$ & $0.05$ & $0.10$ \\
% \midrule
% En$\rightarrow$De & 36.86 & 38.86 & 38.84 \\
% De$\rightarrow$En & 41.10 & 42.46 & 42.40 \\
\midrule
Avg. & 38.98 & 40.66 & 40.62 \\
\bottomrule
\end{tabular}
\caption{Ablation on the router jitter noise ($\eta$) for the \textsc{XL} variant on the multi-domain benchmark.}
\label{tab:abl_jitter}
\end{table}

\begin{figure}[!ht]
    \centering
    \includegraphics[width=0.98\linewidth]{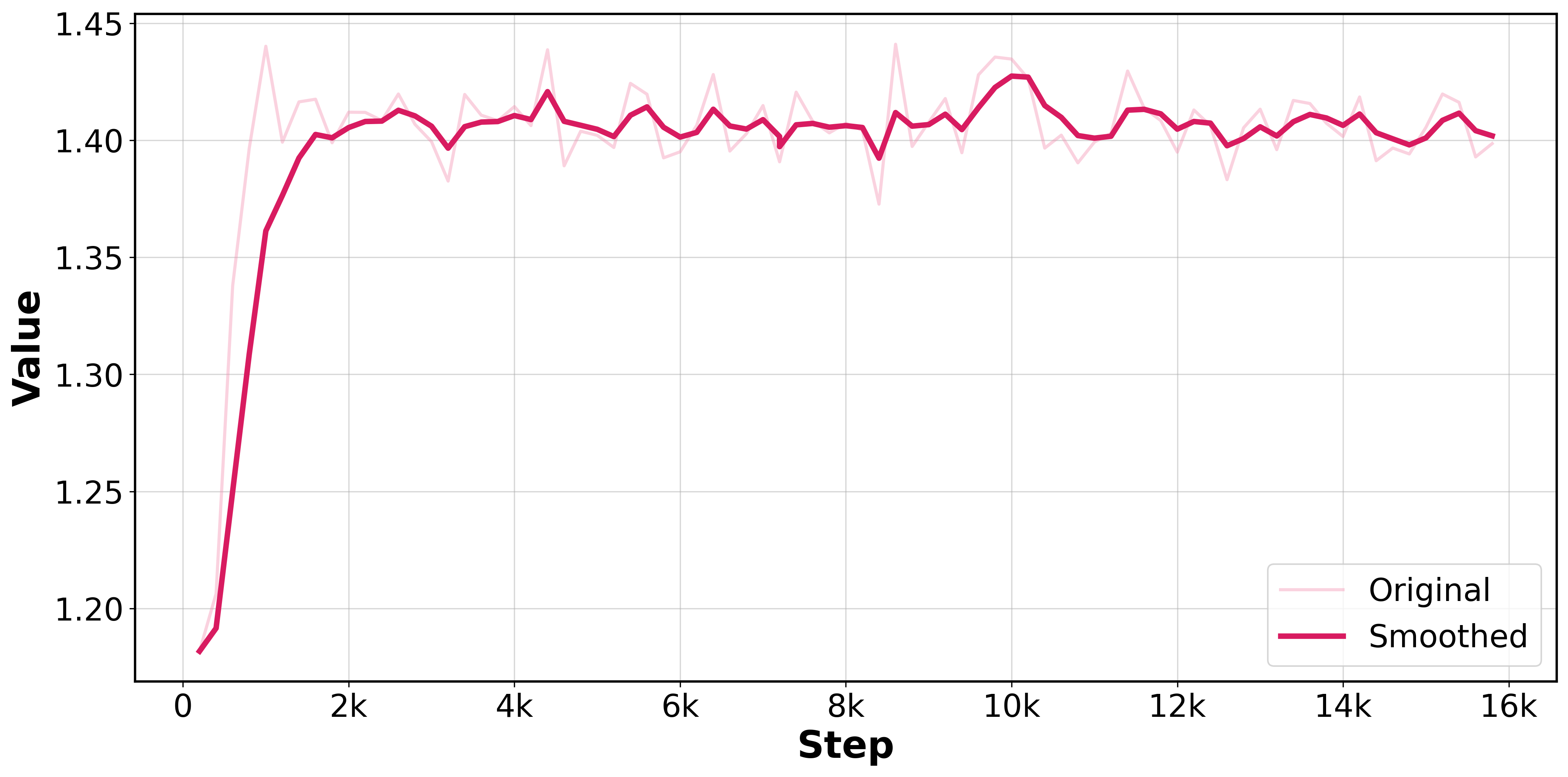}
    \caption{Average active experts per token during training.}
    \label{fig:active_experts_per_token}
\end{figure}

\begin{figure}[!ht]
    \centering
    \includegraphics[width=0.98\linewidth]{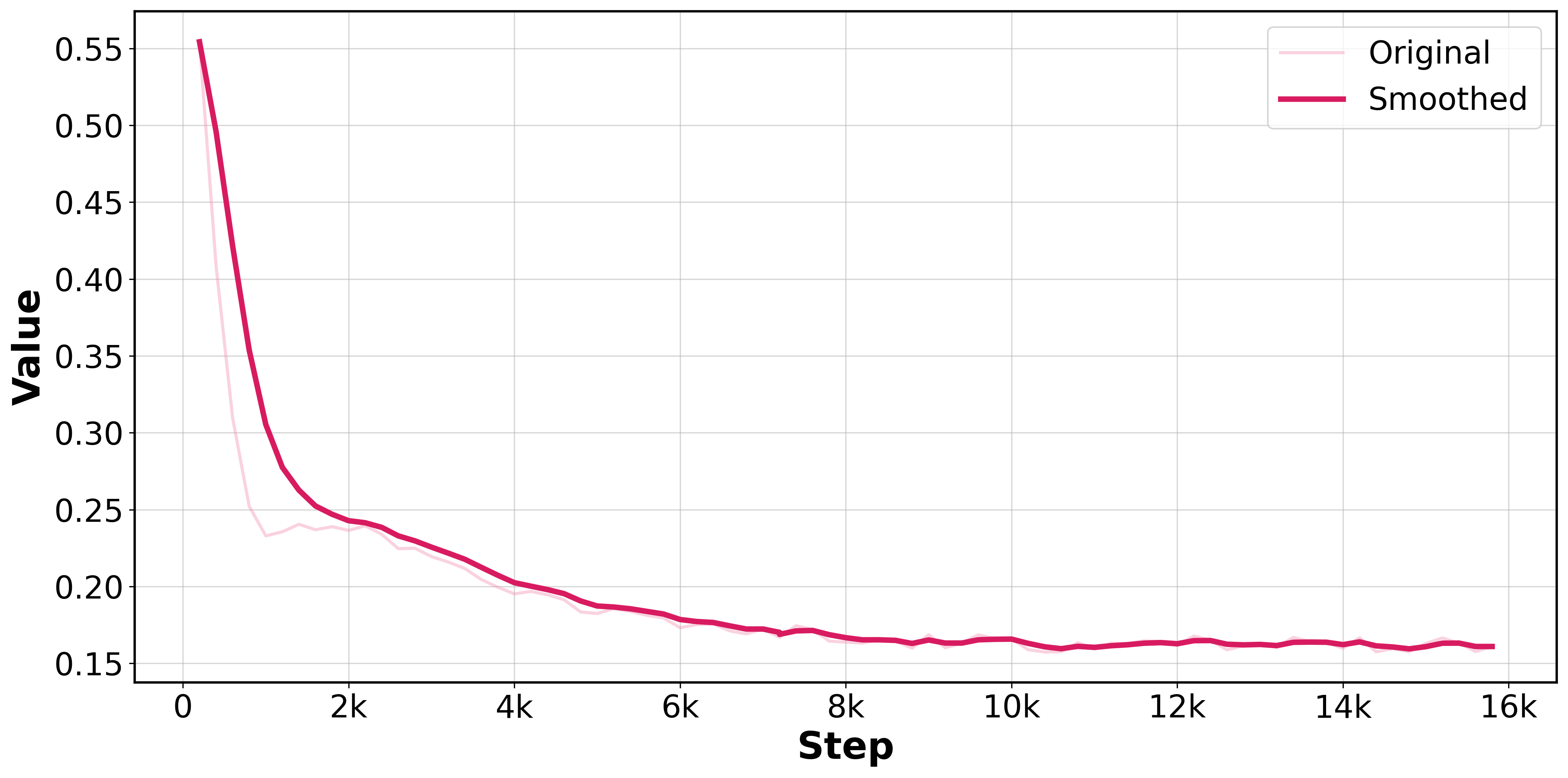}
    \caption{Learnable threshold $\tau$ over the course of training.}
    \label{fig:learnable_tau}
\end{figure}

\begin{figure*}[!ht]
    \centering
    \begin{tabular}{@{}c@{\hspace{0.02\textwidth}}c@{}}
        % Column Titles
        \textbf{Encoder} & \textbf{Decoder} \\[0.5em]
        
        % Row 1: D=256
        \multicolumn{2}{c}{\small$D=256$} \\
        \includegraphics[width=0.48\textwidth]{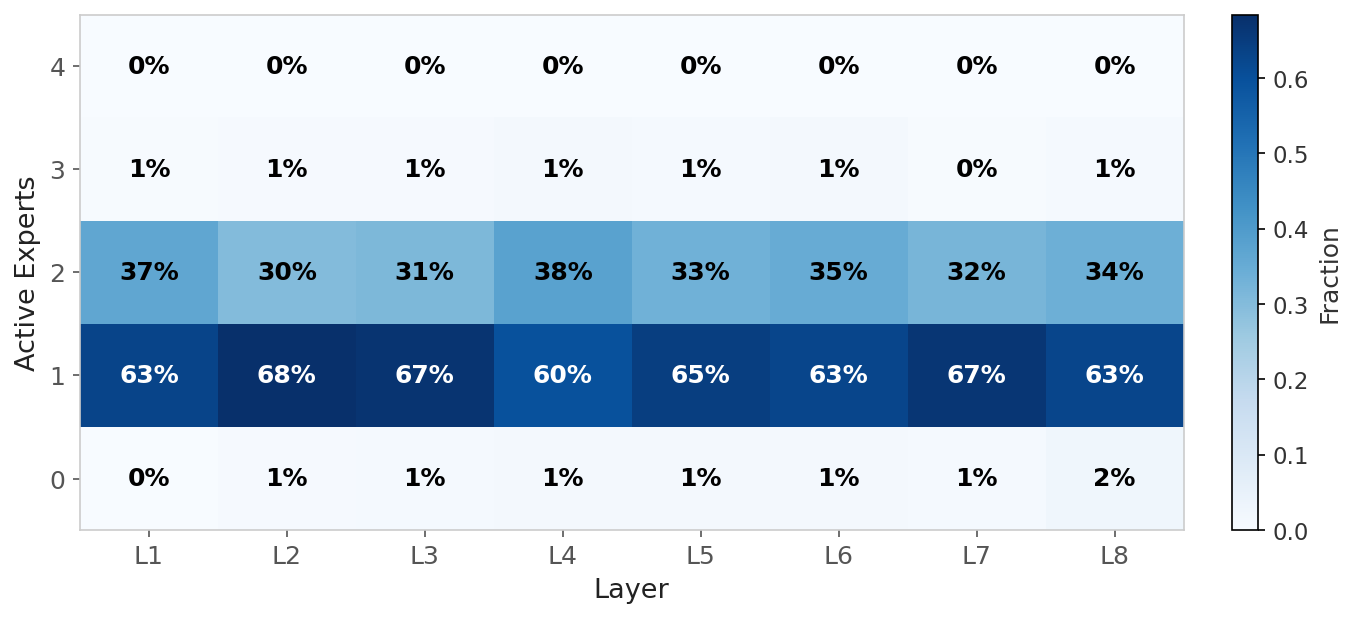} &
        \includegraphics[width=0.48\textwidth]{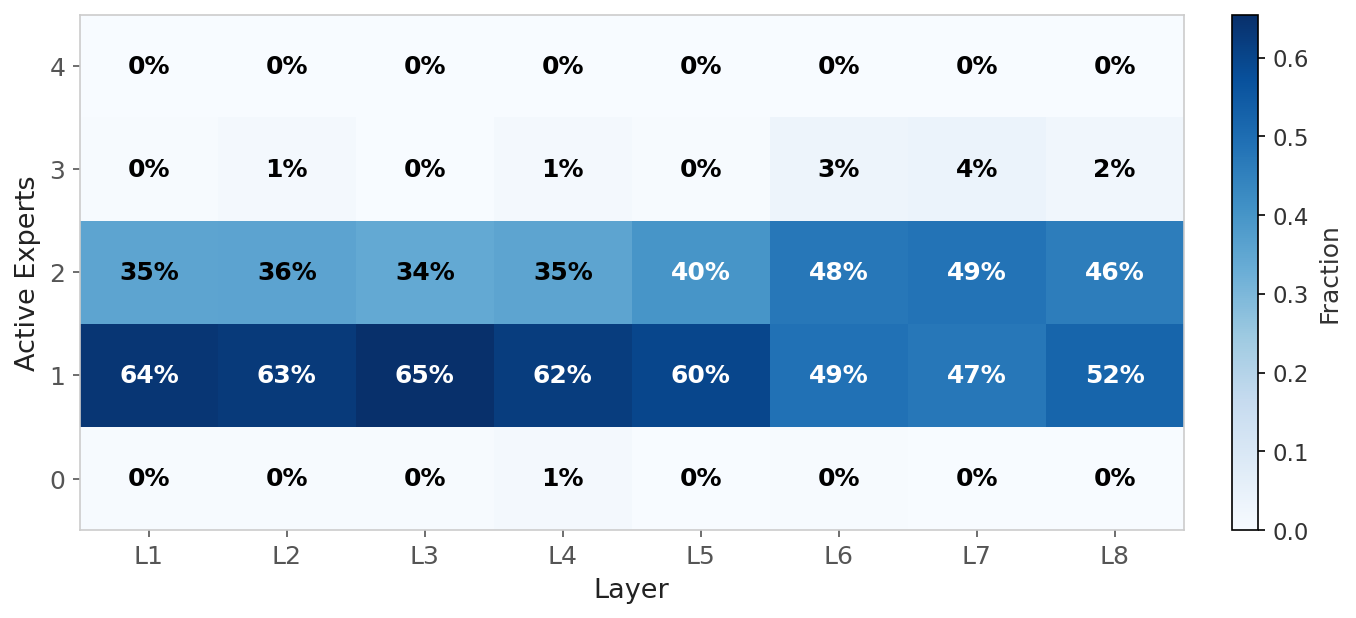} \\[1em]

        % Row 2: D=512
        \multicolumn{2}{c}{\small$D=512$} \\
        \includegraphics[width=0.48\textwidth]{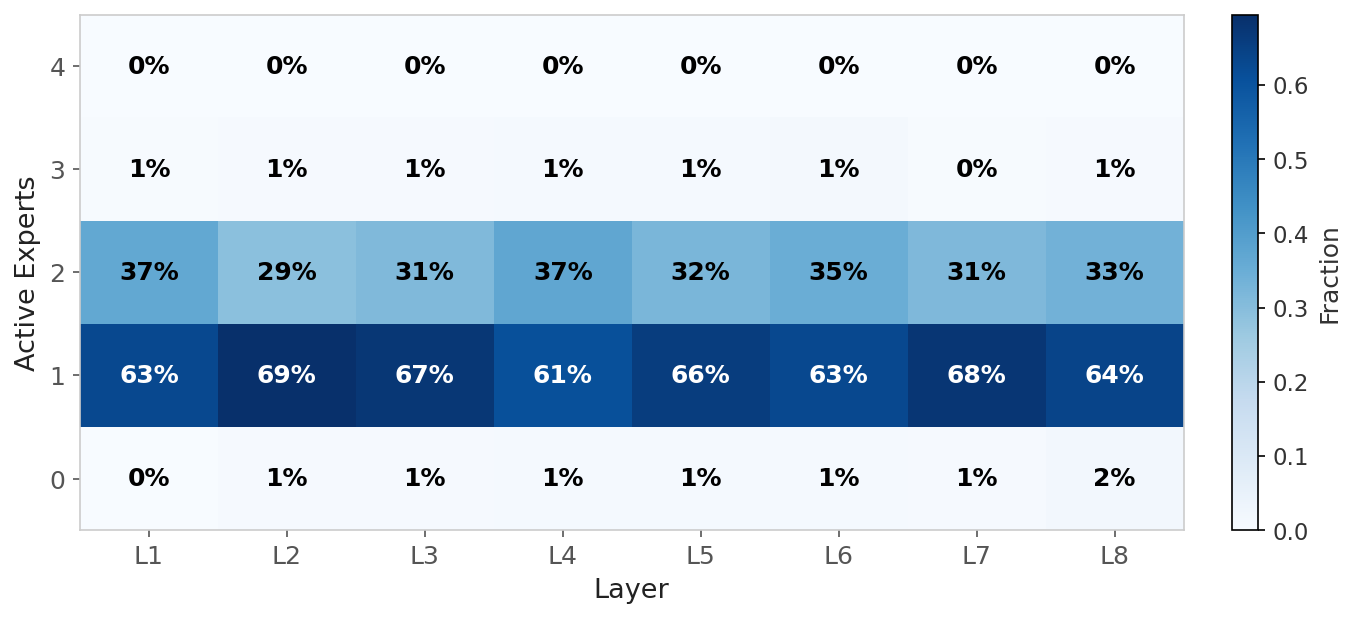} &
        \includegraphics[width=0.48\textwidth]{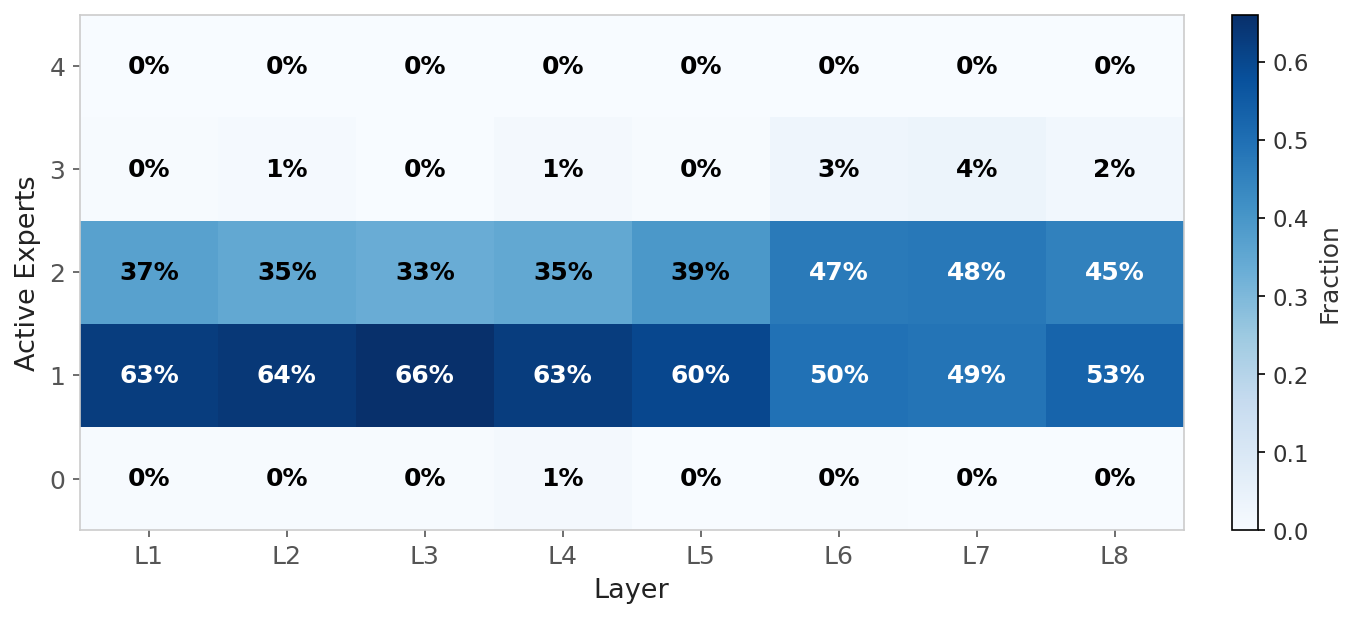} \\[1em]

        % Row 3: D=1024
        \multicolumn{2}{c}{\small$D=1024$} \\
        \includegraphics[width=0.48\textwidth]{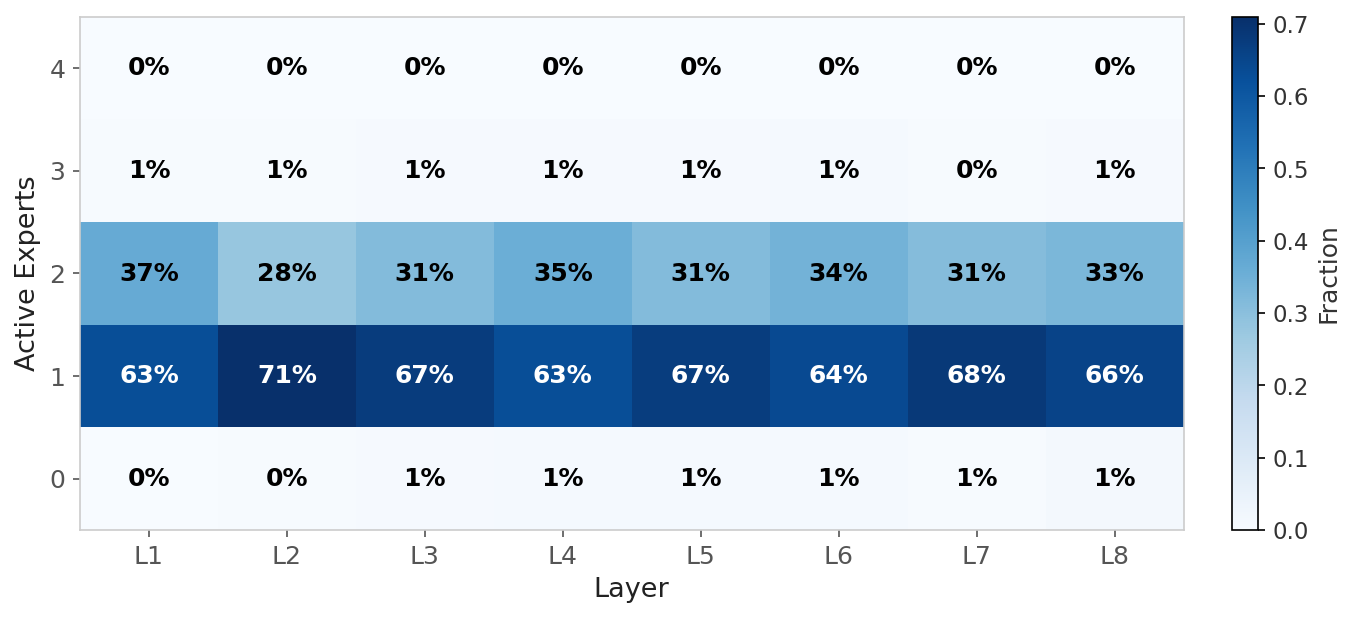} &
        \includegraphics[width=0.48\textwidth]{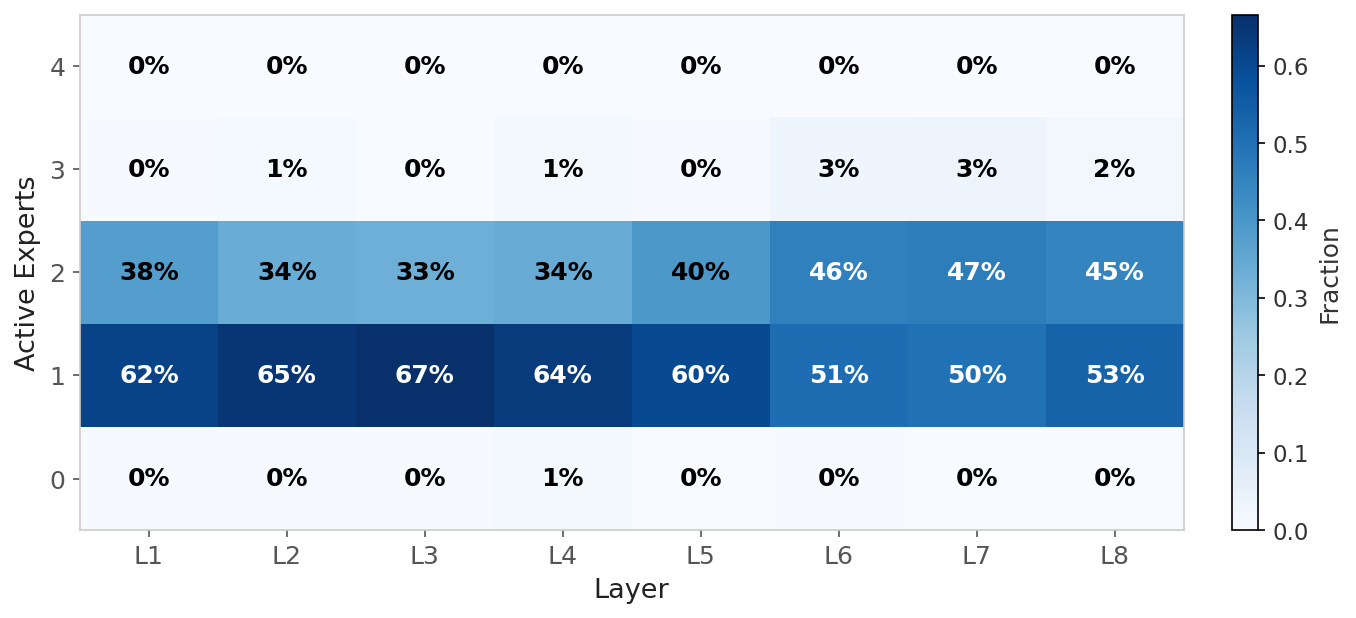} \\[1em]

        % Row 4: D=2048
        \multicolumn{2}{c}{\small$D=2048$} \\
        \includegraphics[width=0.48\textwidth]{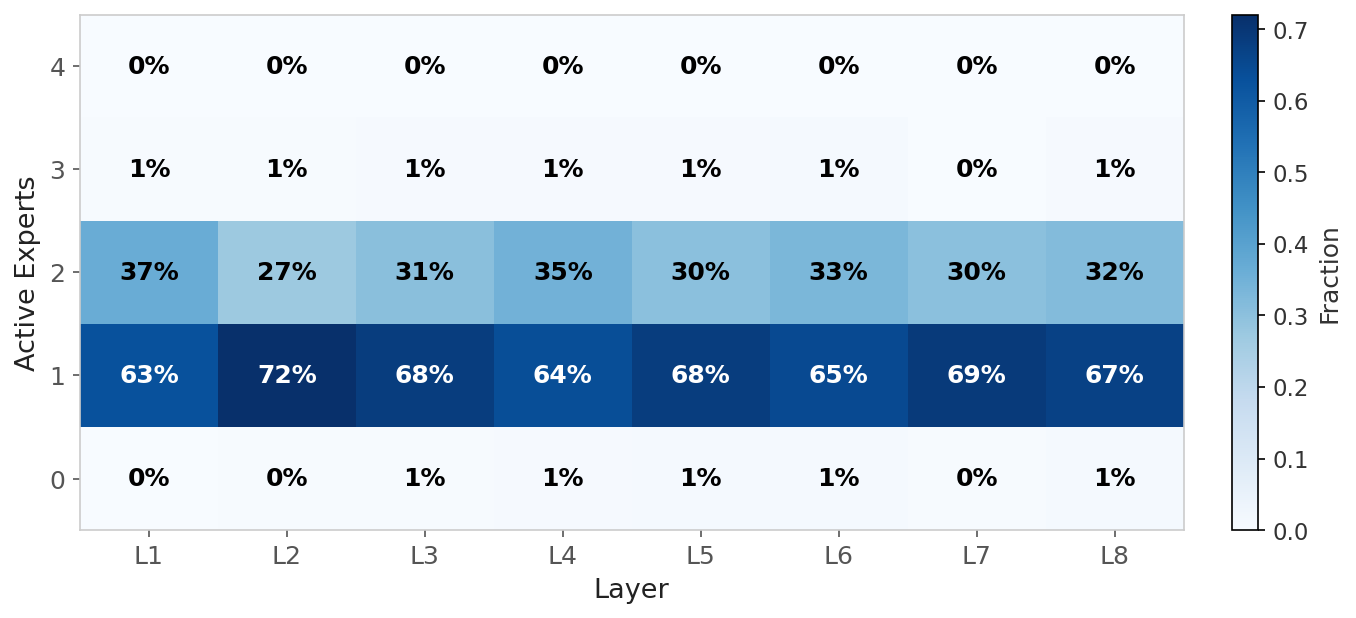} &
        \includegraphics[width=0.48\textwidth]{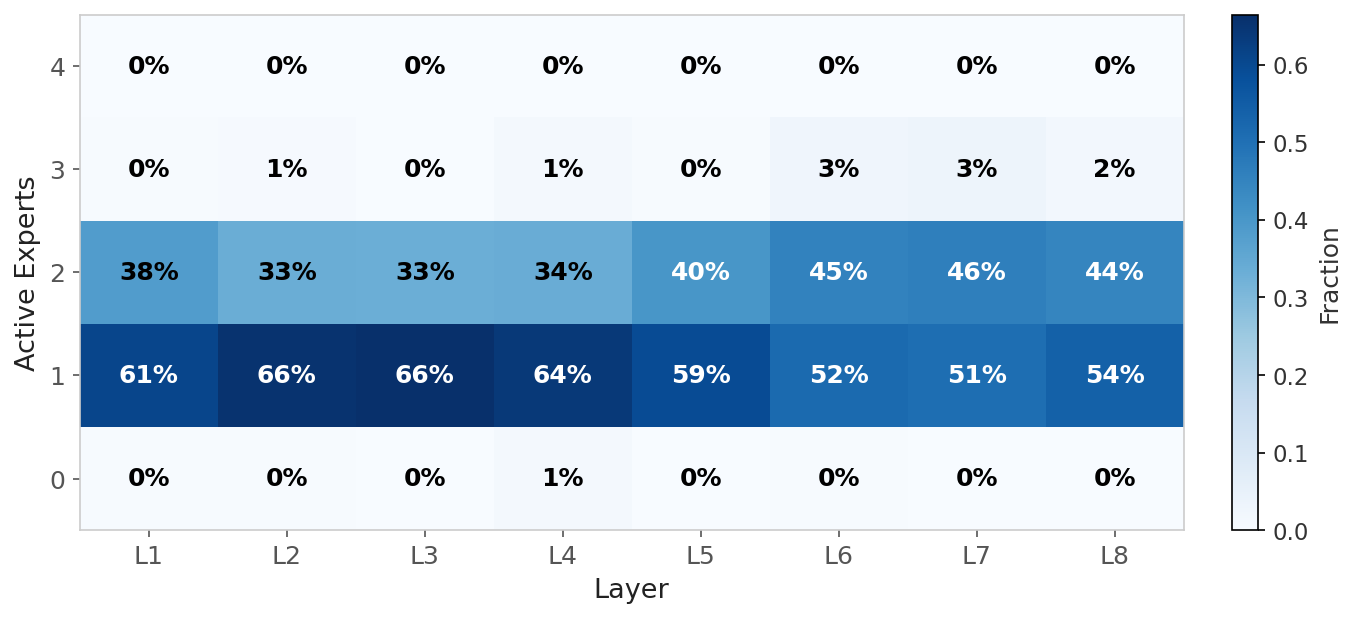} \\
    \end{tabular}
    \caption{Percentage of number of active experts per layer across different elastic dimensions on the multi-domain benchmark.}
    \label{fig:layer_active_params}
\end{figure*}

\begin{table}[!ht]
\small
\centering
\begin{tabular}{l cccc}
\toprule
& \multicolumn{4}{c}{\textbf{En$\leftrightarrow$De}} \\
\cmidrule(lr){2-5}
Setting & XL & L & M & S \\
\midrule
$\lambda_{dc}=0.1$ & \textbf{40.98} & \underline{40.26} & \underline{39.44} & \underline{38.34} \\
$\lambda_{dc}=1.0$ & \underline{40.75} & \textbf{40.28} & \textbf{39.56} & \textbf{38.65} \\
$\lambda_{dc}=5.0$ & 38.33 & 38.23 & 37.92 & 37.44 \\
\bottomrule
\end{tabular}
\caption{Ablation on the Dual-Consistency loss weight ($\lambda_{dc}$) on the multi-domain benchmark, averaged over both translation directions.}
\label{tab:ablation_dc_weight}
\end{table}

\begin{table}[!ht]
\small
\centering
\begin{tabular}{lccc}
\toprule
\multirow{2}{*}{Method} & \multicolumn{3}{c}{\textbf{Target Density ($\rho$)}} \\
\cmidrule(lr){2-4}
& $\rho=1.0$ & $\rho=1.25$ & $\rho=1.5$ \\
\midrule
EDWARD-NMT-S  & 1.18 & 1.39 & 1.69 \\
EDWARD-NMT-M  & 1.17 & 1.38 & 1.67 \\
EDWARD-NMT-L  & 1.16 & 1.37 & 1.66 \\
EDWARD-NMT-XL & 1.15 & 1.37 & 1.65 \\
\bottomrule
\end{tabular}
\caption{Average active routed experts per token on the Multi-domain benchmark across different target density ($\rho$) settings. Values are averaged across both translation directions and all domains.}
\label{tab:density_settings}
\end{table}

Other additional ablations on hyperparameters. We train only the largest variant \textsc{EDWARD-NMT-XL} ($D=2048$), without DC Loss. We report the evaluation result of the hyperparameters in Table~\ref{tab:abl_density}, Table~\ref{tab:abl_tau_min} and Table~\ref{tab:abl_jitter}.

We report the active experts and the dynamic threshold of the model during training in Figure~\ref{fig:active_experts_per_token} and Figure~\ref{fig:learnable_tau}. The number of active experts per layer in encoder and decoder of each variant is reported in Figure~\ref{fig:layer_active_params}.

Table~\ref{tab:ablation_dc_weight} shows the impact of the consistency weight $\lambda_{dc}$. A smaller weight ($\lambda_{dc}=0.1$) favors the full-capacity \textsc{XL} model but yields suboptimal smaller variants due to insufficient knowledge transfer. Conversely, an excessively large weight ($\lambda_{dc}=5.0$) over-regularizes and degrades all scales. Our default $\lambda_{dc}=1.0$ provides the optimal trade-off, maximizing the quality of the deployment-critical \textsc{L, M,} and \textsc{S} sub-networks.

Table~\ref{tab:density_settings} validates our sparsity constraint. By adjusting the target density $\rho$, the empirical active experts scale predictably across all capacities (e.g., from $\sim$1.15 experts at $\rho=1.0$ up to $\sim$1.69 at $\rho=1.5$). This confirms that our auxiliary loss successfully grants fine-grained, predictable control over the inference compute budget without requiring any architectural modifications.

\end{document}